%% file: HRDFuse.tex
\documentclass[10pt,twocolumn,letterpaper]{article}

\usepackage{cvpr}              

\usepackage{graphicx}
\usepackage{amsmath}
\usepackage{amssymb}
\usepackage{booktabs}
\usepackage{graphics}
\usepackage{epstopdf}
\usepackage{color}
\usepackage{multirow}

\definecolor{hollywoodcerise}{rgb}{0.96, 0.0, 0.63}
\definecolor{lasallegreen}{rgb}{0.03, 0.47, 0.19}
\definecolor{hanpurple}{rgb}{0.32, 0.09, 0.98}
\definecolor{green(pigment)}{rgb}{0.0, 0.65, 0.31}
\usepackage[pagebackref=true,breaklinks=true,letterpaper=true,colorlinks,bookmarks=false]{hyperref}
\hypersetup{colorlinks,linkcolor={red},citecolor={hollywoodcerise},urlcolor={red}}  
\usepackage{listings}
\usepackage[capitalize]{cleveref}
\crefname{section}{Sec.}{Secs.}
\Crefname{section}{Section}{Sections}
\Crefname{table}{Table}{Tables}
\crefname{table}{Tab.}{Tabs.}

\newcommand{\textred}{\textcolor[rgb]{1,0,0}}

\def\cvprPaperID{12398} 
\def\confName{CVPR}
\def\confYear{2023}

\begin{document}

\title{HRDFuse: Monocular 360$^\circ$ Depth Estimation by Collaboratively Learning Holistic-with-Regional Depth Distributions}
\author{Hao Ai$^{1,2}$\thanks{Intern at ARC Lab, Tencent PCG.} \quad Zidong Cao$^{1}$ \quad Yan-Pei Cao$^{2}$ \quad Ying Shan$^{2}$ \quad Lin Wang$^{1}$$^{,3}$\thanks{Corresponding author (e-mail: linwang@ust.hk)}\\
$^{1}$AI Thrust, HKUST(GZ) \quad $^{2}$ARC Lab, Tencent PCG \quad $^{3}$Dept. of CSE, HKUST\\
{\tt\small hai033@connect.hkust-gz.edu.cn, caozidong1996@gmail.com}\\
{\tt\small caoyanpei@gmail.com, yingsshan@tencent.com, linwang@ust.hk}}


\maketitle

\begin{abstract} 
Depth estimation from a monocular 360$^\circ$ image is a burgeoning problem owing to its holistic sensing of a scene.
Recently, some methods, \eg, OmniFusion, have applied the tangent projection (TP) to represent a 360$^\circ$ image and predicted depth values via patch-wise regressions, which are merged to get a depth map with equirectangular projection (ERP) format.
However, these methods suffer from 1) non-trivial process of merging plenty of patches; 2) 
capturing less holistic-with-regional contextual information by directly regressing the depth value of each pixel.
In this paper, we propose a novel framework, \textbf{HRDFuse}, that subtly combines the potential of convolutional neural networks (CNNs) and transformers by collaboratively learning the \textit{holistic} contextual information from the ERP and the \textit{regional} structural information from the TP. 
Firstly, we propose a spatial feature alignment (\textbf{SFA}) module that learns feature similarities between the TP and ERP to aggregate the TP features into a complete ERP feature map in a pixel-wise manner. Secondly, we propose a collaborative depth distribution classification (\textbf{CDDC}) module that learns the \textbf{holistic-with-regional} histograms capturing the ERP and TP depth distributions. As such, the final depth values can be predicted as a linear combination of histogram bin centers. Lastly, we adaptively combine the depth predictions from ERP and TP to obtain the final depth map. Extensive experiments show that our method predicts\textbf{ more smooth and accurate depth} results while achieving \textbf{favorably better} results than the SOTA methods.
\end{abstract}
\vspace{-20pt}
\section*{Multimedia Material}
\vspace{-5pt}
For videos, code, demo and more information, you can visit \href{https://VLIS2022.github.io/HRDFuse/}{https://VLIS2022.github.io/HRDFuse/}

\vspace{-5pt}
\section{Introduction}
\label{sec:intro}
\vspace{-3pt}
The 360$^\circ$ camera is becoming increasingly popular as a 360$^\circ$ image provides holistic sensing of a scene with a wide field of view (FoV)~\cite{Ai2022DeepLF, Zhang2019AN3, Gardner2019DeepPI, Barron2022MipNeRF3U, Wang2021LED2NetM3}. Therefore, the ability to infer the 3D structure of a 360$^\circ$ camera's surroundings has sparked the research for monocular 360$^\circ$ depth estimation~\cite{Wang2020BiFuseM3, Wang2022BiFuseSA, Jiang2021UniFuseUF, Shen2022PanoFormerPT}. Generally, raw 360$^\circ$ images are transmitted into 2D planar representations while preserving the omnidirectional information~\cite{yoon2022spheresr, Coxeter1961IntroductionTG}. Equirectangular projection (ERP) is the most commonly used projection format~\cite{Su2017LearningSC, Yang2019DuLaNetAD} and can provide a complete view of a scene. Cubemap projection (CP)~\cite{Cheng2018CubePF} projects 360$^\circ$ contents into six discontinuous faces of a cube to reduce the distortion; thus, the pre-trained 2D convolutional neural networks (CNNs) can be applied. However, ERP images suffer from severe distortions in the polar regions, while CP patches are hampered by geometric discontinuity and limited FoV.

\begin{figure}[!t]
\centering
\includegraphics[width=0.95 \linewidth,height = 5.2cm]{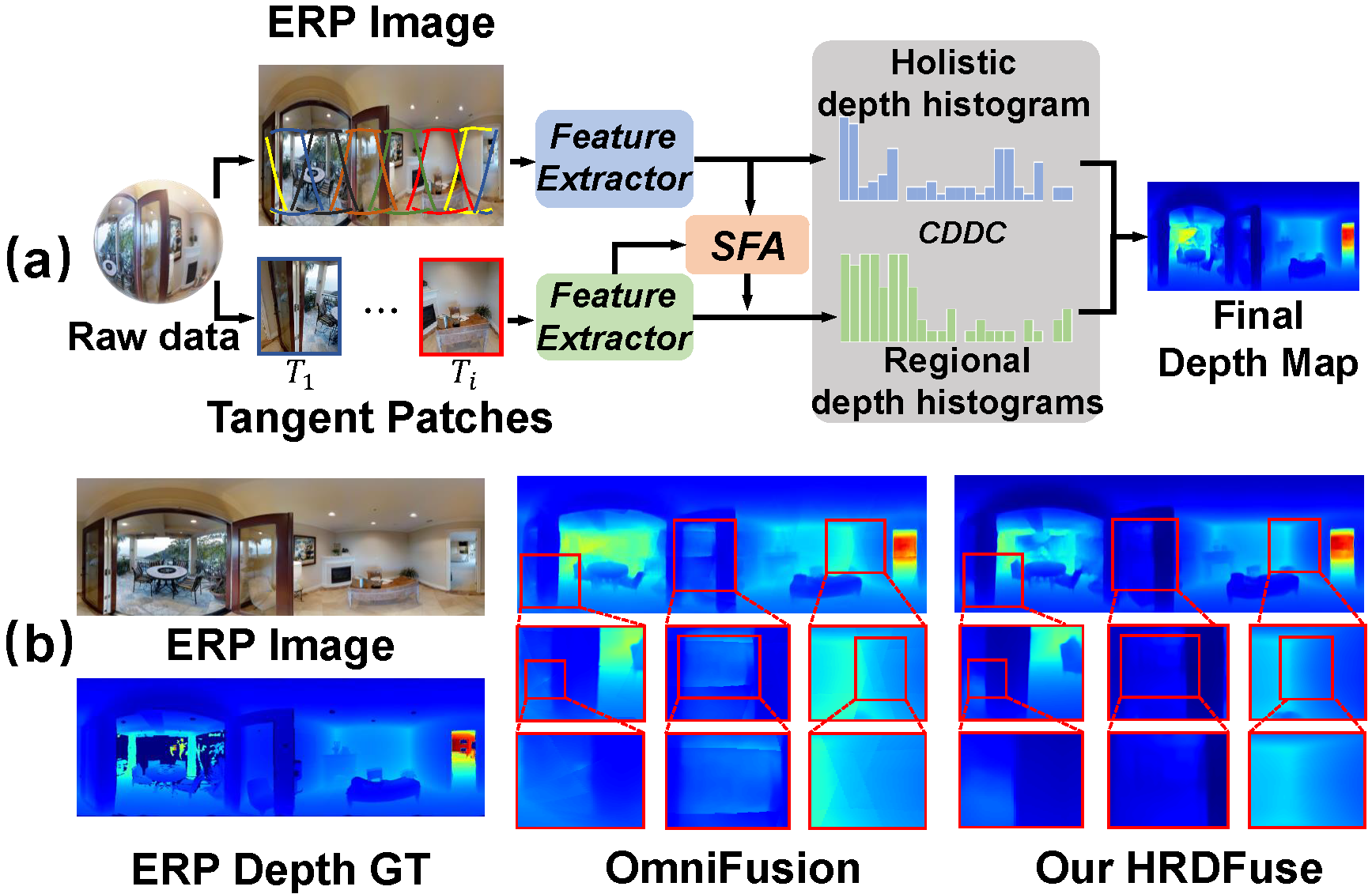}
\vspace{-6pt}
\caption{(a) Our HRDFuse employs the SFA module to align the regional information in discrete TP patches and holistic information in a complete ERP image. The CDDC module is proposed to estimate ERP format depth outputs from both the ERP image and TP patches based on holistic-with-regional depth histograms. (b) Compared with OmniFusion~\cite{Li2022OmniFusion3M}, our depth predictions are more smooth and more accurate.}
\vspace{-15pt}
 \label{fig:coverfig}
\end{figure}

For this reason, some works~\cite{Zioulis2018OmniDepthDD,Zhuang2021ACDNetAC} have proposed distortion-aware convolution filters to tackle the ERP distortion problem for depth estimation.
BiFuse~\cite{Wang2020BiFuseM3} and UniFuse~\cite{Jiang2021UniFuseUF} explore the complementary information from the ERP image and CP patches to predict the depth map. 

\begin{figure}[!t]
    \centering
    \includegraphics[width=1\linewidth]{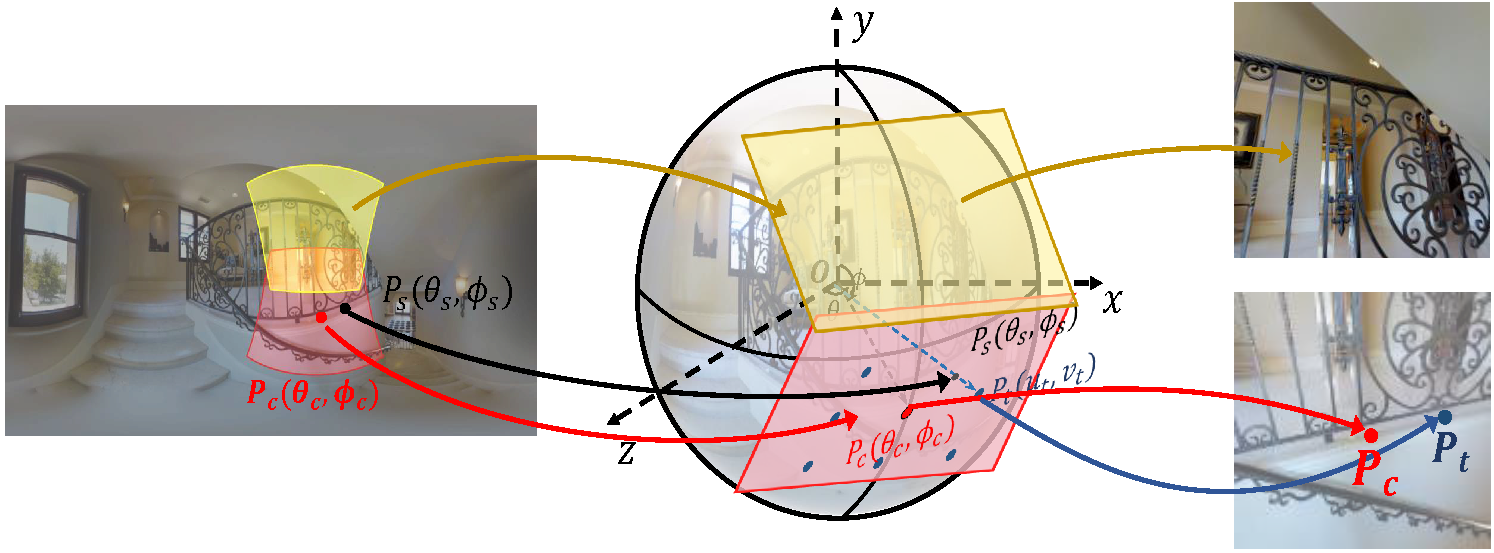}
    \vspace{-12pt}
\caption{Geometric relationship between TP and ERP. Two TP patches are projected from the red area and yellow area.}
    \label{fig:tp-erp}
    \vspace{-16pt}
\end{figure}

Recently, research has shown that it is promising to use tangent projection (TP) because TP patches have less distortion, and many pre-trained CNN models designed for perspective images can be directly applied~\cite{Eder2020TangentIF}. However, there exist unavoidable overlapping areas between two neighbouring TP patches, as can be justified by the geometric relationship in Fig.~\ref{fig:tp-erp}. Therefore, directly re-projecting the results from TP patches into the ERP format is computationally complex. Accordingly, 360MonoDepth~\cite{ReyArea2021360MonoDepthH3} predicts the patch-wise depth maps from a set of TP patches using the state-of-the-art (SOTA) perspective depth estimators, which are aligned and merged to obtain an ERP format depth map. OmniFusion~\cite{Li2022OmniFusion3M} proposes a framework leveraging CNNs and transformers to predict depth maps from the TP inputs and merges these patch-wise predictions to the ERP space based on  geometric prior information to get the final depth output with ERP format. However, these methods suffer from two critical limitations because: 1) geometrically merging a large number of patches is computationally heavy; 2) they ignore the holistic contextual information contained only in the ERP image and directly regress the depth value of each pixel, leading to less smooth and accurate depth estimation results.

\begin{figure*}[t!]
\centering
\includegraphics[width=1\textwidth]{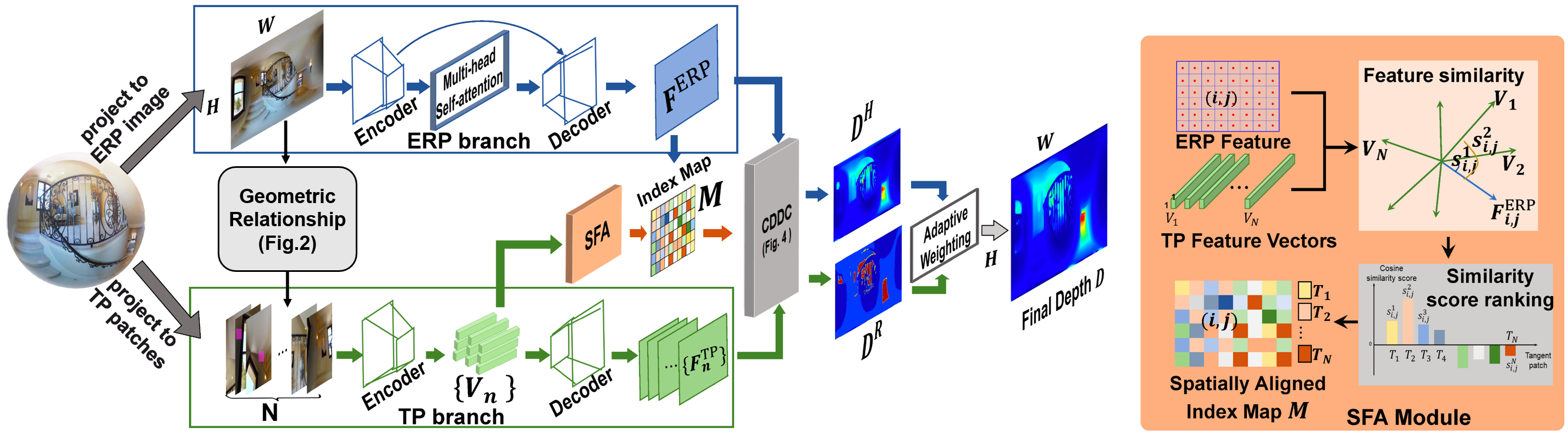}
\vspace{-15pt}
\caption{Overview of our HRDFuse, consisting of three parts: feature extractors for both ERP and TP inputs, spatial feature alignment (SFA) module, and collaborative depth distribution classification (CDDC) module (See Fig.~\ref{fig:cddc} for details).}
\vspace{-15pt}
\label{fig:overview}
\end{figure*}

To tackle these issues, we propose a novel framework, called \textbf{HRDFuse}, that subtly combines the potential of convolutional neural networks (CNNs) and transformers by collaboratively exploring the \textit{\textbf{holistic}} contextual information from the ERP and \textit{\textbf{regional}} structural information from the TP (See Fig.~\ref{fig:coverfig}(a) and Fig.~\ref{fig:overview}). Compared with previous methods, our method achieves more smooth and more accurate depth estimation results while maintaining high efficiency with three key components. Firstly, for each projection, we employ a CNN-based feature extractor
to extract spatially consistent feature maps and a transformer encoder
to learn the depth distribution with long-range feature dependencies. In particular, to efficiently aggregate the individual TP information into an ERP space, we propose a spatial feature alignment (\textbf{SFA}) module to learn a spatially aligned index map based on feature similarities between ERP and TP. With this index map, we can efficiently measure the spatial location of each TP patch in the ERP space and achieve pixel-level fusion of TP information to obtain a smooth output in ERP format.
Secondly, we propose a collaborative depth distribution classification (\textbf{CDDC}) module to learn the \textit{holistic} depth distribution histogram from the ERP image and \textit{regional} depth distribution histograms from the collection of TP patches. Consequently, the pixel-wise depth values can be predicted as a linear combination of histogram bin centers. Lastly, the final result is adaptive fused by two ERP format depth predictions from ERP and TP.

We conduct extensive experiments on three benchmark datasets: Stanford2D3D~\cite{Armeni2017Joint2D}, Matterport3D~\cite{Chang2017Matterport3DLF}, and 3D60~\cite{Zioulis2018OmniDepthDD}. The results show that our method can achieve more smooth and more accurate depth results while favorably surpassing the existing methods by a significant margin on 3D60 and Stanford2D3D datasets (See Fig.~\ref{fig:coverfig} and Tab.~\ref{tab:comparison-to-soat}). In summary, our main contributions are four-fold: (\textbf{I})  We propose HRDFuse that combines the holistic contextual information from the ERP and regional structural information from the TP. (\textbf{II}) We introduce the SFA module to efficiently aggregate the TP features into the ERP format, relieving the need for expensive re-projection operations. (\textbf{III}) We propose the CDDC module to learn the holistic-with-regional depth distributions and estimate the depth value based on the histogram bin centers. 

\vspace{-6pt}
\section{Related Work}
\vspace{-2pt}
\label{sec:related}

\subsection{Monocular 360 Depth Estimation} 
\vspace{-3pt}
\noindent \textbf{ERP-based methods.} To address the spherical distortion in the ERP images, endeavours have been made to leverage the characteristics of convolutional filters. OmniDepth~\cite{Zioulis2018OmniDepthDD} applies row-wise rectangular filters to cope with the distortions in different latitudes, while ACDNet~\cite{Zhuang2021ACDNetAC} leverages a group of dilated convolution filters to rectify the receptive field. Tateno~\etal~\cite{Tateno2018DistortionAwareCF}
explored the standard convolution filters trained with the perspective images, and deformed the shape of sampling grids based on spherical distortion accordingly during the inference. 
SliceNet~\cite{Pintore2021SliceNetDD} partitions an ERP image into vertical slices and directly applies the standard convolutional layers to predict the ERP depth map.

\noindent \textbf{Combination of CP and ERP.} 
BiFuse~\cite{Wang2020BiFuseM3} proposes to bidirectionally fuse the ERP and CP features at both encoding and decoding stages. By contrast, UniFuse~\cite{Jiang2021UniFuseUF} fuses the features only at the encoding stage as it is argued that ERP features are more important for final ERP format depth prediction. Differently,~\cite{Bai2022GLPanoDepthGP} employs CNNs to extract ERP features and a transformer block~\cite{AlexeyDosovitskiy2020AnII} to extract CP features, which are fused to predict the final depth map. Recently, M$^3$PT~\cite{Yan2022MultiModalMP} introduces the shared random masks to process the ERP panoramas and CP depth patches simultaneously and combines the RGB information with sparse depth information to achieve panoramic depth completion.

\noindent \textbf{TP-based methods.}
TP is recently shown to suffer less from distortion (See Fig.~\ref{fig:tp-erp}), and the pre-trained CNN models designed for perspective images can be directly applied~\cite{Coors2018SphereNetLS}. Accordingly, 360MonoDepth~\cite{ReyArea2021360MonoDepthH3} and OmniFusion~\cite{Li2022OmniFusion3M} build their frameworks based on the TP patches. Concurrently, PanoFormer~\cite{Shen2022PanoFormerPT} proposes a transformer-based architecture to process the TP patches as the tokens for depth estimation. Different from \cite{ReyArea2021360MonoDepthH3,Li2022OmniFusion3M}, PanoFormer designs a handcrafted sampling method to form a tangent patch by sampling eight most relevant tokens for each central token on the ERP domain rather than dividing the ERP images into TP patches. However, PanoFormer is limited by the inference speed (See Sec.~\ref{abl_study}), and ignores the 2D structure and spatial local information within each patch~\cite{Guo2022CMTCN}.
For more details, we refer readers to a recent survey~\cite{Ai2022DeepLF}. 
In comparison, our HRDFuse combines the potential of CNNs and transformers by collaboratively learning the holistic contextual information from the ERP image and regional structural information from the TP patches.

\vspace{-5pt}
\subsection{Distribution-based Planar Depth Estimation}
\vspace{-3pt}
Many methods estimate depth by directly regressing the depth values; however, they suffer from slow convergence, and deficiency of global analysis~\cite{IroLaina2016DeeperDP,Lee2021PatchWiseAN}. For this reason, ~\cite{Fu2018DeepOR} discretized the depth range into several pre-determined intervals and recast depth prediction as an ordinal regression problem, which accounts the depth distributions depending on the located intervals. Adabins~\cite{Bhat2021AdaBinsDE} divides the depth range into many adaptive bins whose widths are computed from the scene information, and the depth values are a linear combination of the bin centers, showing better performance over previous methods. Our HRDFuse is the \textit{first} to explore the idea of depth distribution classification for 360$^\circ$ depth estimation. The proposed CDDC module learns the holistic depth distribution histograms from the ERP image and regional depth distribution histograms from the collection of TP patches. As such, the final depth values are predicted as a linear combination of bin centers.
\vspace{-5pt}
\subsection{Vision Transformer} 
\vspace{-5pt}
Transformers are capable of modeling the long-range dependencies for computer vision tasks~\cite{AlexeyDosovitskiy2020AnII}. Recently, it has been shown that the combination of convolutional operations and self-attention mechanisms further enhance the representation learning. 
For instance, DeiT~\cite{Touvron2021TrainingDI} employs a CNN as the teacher model to distill the tokens to the transformer, while DETR~\cite{Carion2020EndtoEndOD} models the global relationship via serially feeding the features extracted by CNNs to the transformer encoder-decoder. Moreover,  some works,\eg,~\cite{Peng2021ConformerLF,Chen2021MobileFormerBM} attempted to concurrently fuse the features from CNNs and transformers. Our HRDFuse framework is also built based on the combination of CNNs and transformers; however, it shares a different spirit as we focus on ensuring network efficiency. Thus, we extract the high-resolution feature maps using a CNN-based encoder-decoder and feed them to a smaller transformer encoder~\cite{AlexeyDosovitskiy2020AnII} to estimate distributions.

\vspace{-5pt}
\section{Methodology}
\label{method}
\vspace{-2pt}
\subsection{Overview}
\vspace{-2pt}
\label{sec:overview}
As depicted in Fig.~\ref{fig:overview}, to exploit the complementary information from holistic context and regional structure, our framework simultaneously takes two projections of a 360$^\circ$ image, an ERP image and $N$ TP patches, as inputs. For the ERP branch (See Fig.~\ref{fig:overview} Top), an ERP image with the resolution of $H\times W$ is fed into a feature extractor, comprised of an encoder-decoder block, to produce a decoded ERP feature map $F^{\mathrm{ERP}}$. For the TP branch (See Fig.~\ref{fig:overview} Bottom), $N$ TP patches are first obtained with gnomonic projection from the same sphere~\cite{Eder2020TangentIF}. This indicates that the feature distributions of TP branch are closely correlated with those of the ERP branch, similar to ERP-to-CP (E2C) or C2E feature transform in \cite{Wang2020BiFuseM3}. Then, these TP patches are passed through the TP feature extractor to obtain 1-D patch feature vectors $\left\{V_{n}, n=1,\dots,N\right\}$, which are passed through the TP decoder to obtain the TP feature maps $\left\{F^{\mathrm{TP}}_{n},n=1,\dots, N\right\}$. 

To determine and align the spatial location of each TP patch in the ERP space and avoid complex geometric fusion for overlapping areas between neighboring TP patches, we propose the spatial feature alignment (SFA) module (Fig.~\ref{fig:overview}) to learn feature correspondences between pixel vectors in the ERP feature map $F^{\mathrm{ERP}}$ and patch feature vectors $\left\{V_n\right\}$. This way, we can obtain the spatially aligned index map $M$, recording the location of each patch in the ERP space.

Next, the index map $M$, ERP feature map $F^{\mathrm{ERP}}$, and TP feature maps $\left\{F_n^{\mathrm{TP}}\right\}$ are fed into the proposed collaborative depth distribution classification (CDDC) module that accordingly outputs two ERP format depth predictions (See Fig.~\ref{fig:cddc}).
In principle, the CDDC module first learns holistic-with-regional histograms to simultaneously capture depth distributions from the ERP image and a set of TP patches. Consequently, the depth distributions are then converted to depth values through a linear combination of bin centers.
Lastly, the two depth predictions from the CDDC module are adaptively fused to output the final depth result. We now describe these modules in detail.

\vspace{-3pt}
\subsection{Feature Extraction}
\vspace{-3pt}
\label{sec:fe}

Overall, taking the ERP image and a collection of TP patches as inputs, the feature extractor of the ERP branch outputs the decoded feature map $F^{\mathrm{ERP}}$, and the feature extractor of the TP branch produces encoded patch feature vectors $\left\{V_n\right\}$ and decoded TP feature maps $\left\{F_n^{\mathrm{TP}}\right\}$. 

Specifically, for the ERP branch (Fig.~\ref{fig:overview} Top), we design the feature extractor with an encoder-decoder network, following the design of OmniFusion~\cite{Li2022OmniFusion3M}. It consists of an encoder built with the pre-trained ResNet34~\cite{He2016DeepRL}, a multi-head self-attention block~\cite{vaswani2017attention}, and a decoder with commonly used up-sampling blocks. This way, we obtain the decoded feature map $F^{\mathrm{ERP}}$.

For the TP branch, we first sample TP patches from the sphere via gnomonic projection~\cite{Ai2022DeepLF,Eder2020TangentIF}. \textit{The details can be found in the suppl. material.} Secondly, we feed the patches simultaneously into the feature extractor, similar to the ERP branch but without the multi-head self-attention block, which helps to maintain the independence of each patch feature vector for spatial feature alignment. 
As such, we extract the patch feature vectors $\left\{V_{n}\right\}$ through the encoder and obtain the decoded patch feature maps $\left\{F_n^{\mathrm{TP}}\right\}$. 
The resolutions of the ERP feature map $F^{\mathrm{ERP}}$ and TP feature maps $\left\{F_n^{\mathrm{TP}}\right\}$ are set to half of the corresponding input resolutions for efficiency.

\vspace{-3pt}
\subsection{Spatial Feature Alignment}
\label{sec:SFA}
With ERP feature map $F^{\mathrm{ERP}}$ and patch feature vectors $\left\{V_{n}\right\}$, our SFA module outputs the spatially aligned index map $M$. \textit{It determines the spatial relations between the individual TP patches and pixel positions in the complete ERP space according to the feature similarity score ranking (See Fig.~\ref{fig:overview}) and can be applied to achieve smooth pixel-wise fusion of individual TP information.}
Existing works aggregate the discrete TP information into the complete ERP space via geometric fusion~\cite{Li2022OmniFusion3M,ReyArea2021360MonoDepthH3}.
However, they are less capable of predicting smooth equirectangular depth outputs without holistic contextual information. For instance, as shown in Fig.~\ref{fig:coverfig}(b), depth predictions in OmniFusion suffer from severe artifacts along the edges of the merged regions. For this reason, we propose the SFA module to measure, rank, and record the pixel-wise similarities between the ERP feature map $F^{\mathrm{ERP}}$ and patch feature vectors $\left\{V_{n}\right\}$. The pixel-wise similarity can be formulated as:
\begin{equation}
\setlength\abovedisplayskip{2pt}
\setlength\abovedisplayskip{2pt}
s_{(i,j), k} = \frac{\overrightarrow{F^{\mathrm{ERP}}(i,j)} \cdot \overrightarrow{V_k}}{\left \|F^{\mathrm{ERP}}(i,j)  \right \|\left \| V_k\right \|},
\end{equation}
where $(i,j)$ is the coordinate of a pixel in the ERP feature map and $k$ is the TP patch index. As depicted in Fig.~\ref{fig:overview}, for each feature vector $F^{\mathrm{ERP}}(i,j)$ in the ERP feature map, our SFA module calculates the cosine similarity score $s_{(i,j),k}$ between $F^{\mathrm{ERP}}(i,j)$ and each patch feature vector $V_{k}$. Then, it ranks the scores, and selects the $m$-th patch that satisfies:
\begin{equation}
\resizebox{0.38\hsize}{!}{$
    m = \mathop{\arg\max}\limits_{k}{s_{(i,j), k}}, 
$}
\end{equation}
and records the index $m$ of the pixel location $(i,j)$ on the spatially aligned index map $M$. For convenience, we extend each index into an N-dimension one-hot vector and transform the resolution size of index map $M$ to $h_e \times w_e$, where $(h_e, w_e)$ is the resolution size of ERP feature map $F^{\mathrm{ERP}}$. Note that this spatially aligned index map is produced with the guidance of the holistic contextual information only contained in the ERP image. With this index map, we can efficiently aggregate the TP features into an ERP format feature map while maintaining spatial consistency.

\begin{figure}[!t]
    \centering\includegraphics[width=0.98\linewidth,height=7cm]{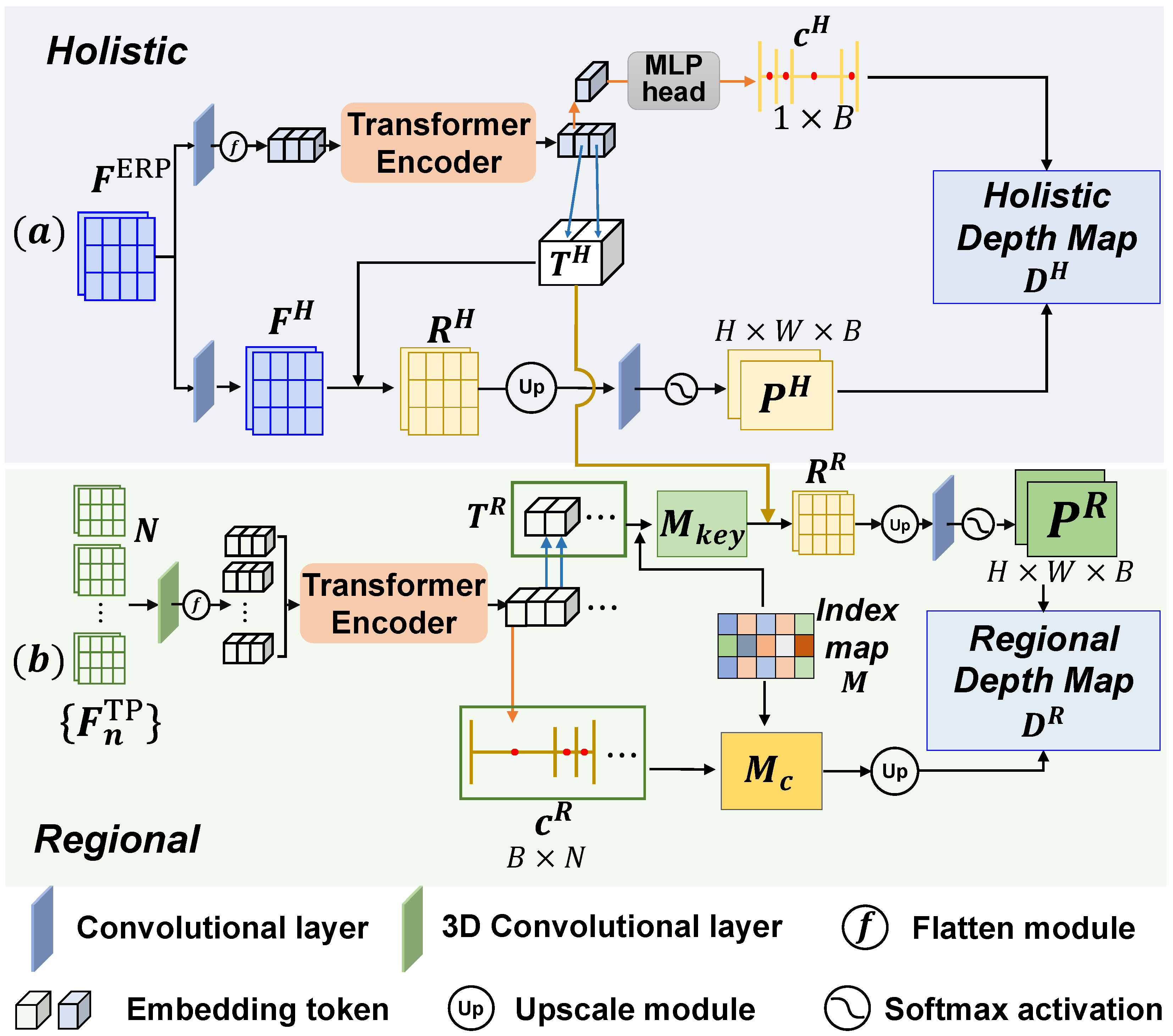}
\caption{Overview of the CDDC module with two steps: depth distribution histogram classification, and depth prediction combination based on the range attention maps.}
    \label{fig:cddc}
    \vspace{-8pt}
\end{figure}

\begin{figure*}[t!]
    \centering
    \includegraphics[width=.96\textwidth]{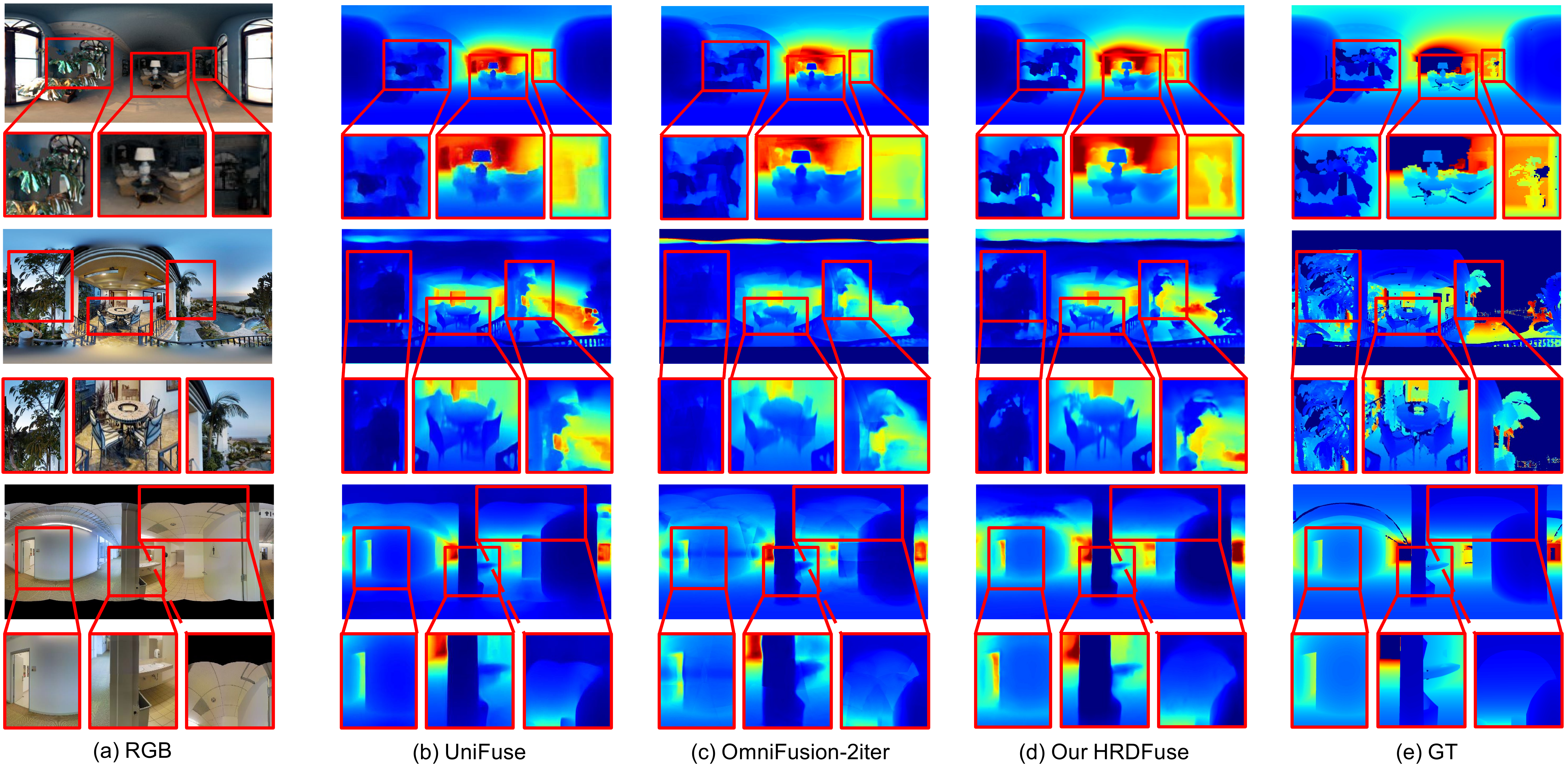}
     \vspace{-5pt}
\caption{Qualitative results on 3D60 (top), Matterport3D (middle) and Stanford2D3D (bottom).}
    \label{fig:compare}
     \vspace{-15pt}
\end{figure*}
\vspace{-3pt}
\subsection{Collaborative Depth Distribution Classification}
\label{sec:CDDC}

The proposed CDDC module replaces the pixel-wise depth value regression with depth distribution classification, inspired by the works for perspective images~\cite{Bhat2021AdaBinsDE,Fu2018DeepOR}. Importantly, to fully exploit the complete view in the ERP image and structural details in the less-distorted TP patches, we marry the potential of CNNs and transformers to learn the holistic-with-regional histograms capturing the ERP and TP depth distributions simultaneously. 
In the following, we introduce our CDDC module in three parts: generic depth distribution classification, depth prediction based on the holistic depth distribution, and depth prediction based on the regional depth distributions.

\noindent \textbf{Generic depth distribution classification.} 
Following previous works~\cite{Bhat2021AdaBinsDE,Fu2018DeepOR}, given an extracted feature map $F \in \mathbb{R}^{{H} \times W \times C_{in}}$ (\eg, $F^{\mathrm{ERP}}$ in Fig~\ref{fig:cddc}(a)), a sequence of embedding tokens $T_{in}$ is obtained from $F$ by a convolutional layer followed by a spatially flattening module. A transformer encoder then encodes the embedding tokens $T_{in}$, producing processed tokens $T_{out}$. Note that the processed tokens $T_{out}$ now benefit from the global context and thus can accurately capture the depth distribution. Then the first token $T_{out}[0]$ from $T_{out}$ is selected to predict the bin centers $\mathbf{c}$ of depth distribution histograms (\eg, $\mathbf{c}^H$ in Fig~\ref{fig:cddc}(a)) as:
\begin{equation}
\setlength{\abovedisplayskip}{2pt}
\setlength{\belowdisplayskip}{2pt}
    \mathbf{c}_i = D_{min} +  (\mathbf{w}_i/2+\sum_{j=1}^{i-1}\mathbf{w}_j),
\end{equation}
\begin{equation}
\setlength{\abovedisplayskip}{2pt}
\setlength{\belowdisplayskip}{2pt}
    \mathbf{w}_i= (D_{max}-D_{min})\frac{(\mathrm{mlp}(T_{out}[0]))_i+ \epsilon}{\sum_{j=1}^{B}{(\mathrm{mlp}(T_{out}[0]))_j + \epsilon}},
\end{equation}
where $i,j=1,\dots,B$, $\mathbf{w}$ is the bin widths of the distribution histogram,  $\mathrm{mlp}$ denotes a multi-layer perceptron (MLP) head with a ReLU activation, $(D_{min}, D_{max})$ is the depth range of the dataset, $B$ denotes the number of depth distribution bins,  and $\epsilon$ is a small constant to ensure that each value of $\mathbf{w}$ is positive. 
Finally, the bin centers $\mathbf{c}$ are linearly blended with a probability score map $P$ (\eg, $P^H$ in the Fig~\ref{fig:cddc}(a)) to predict the depth value at each pixel $(i,j)$:
\begin{equation}
\resizebox{0.5\hsize}{!}{$
    D(i,j) = \sum_{b=1}^{B} P(i,j)_b \cdot \mathbf{c}_b.
    $}
\label{eq_depth}
\end{equation}


\noindent \textbf{Holistic distribution-based depth prediction.} As depicted in Fig.~\ref{fig:cddc}(a), we follow the process of generic depth distribution classification to predict the holistic depth bin centers $\mathbf{c}^H$. 
We then perform the following steps to obtain the holistic probability score map $P^H$. First, we select a part of processed tokens, which are the output of the transformer encoder and contain global context, as the ``query'' embedding $T^H$. At the same time, we encode a spatially consistent feature map $F^H$ containing local pixel-wise information as the ``keymap''. Next, we calculate the dot-production between the query $T^H$ and pixel features in $F^H$ to obtain a range attention map $R^H$. This range attention map $R^H$ thus contains global context and is spatially aligned with the ERP feature map. Then $R^H$ is passed through a $1\times1$ convolutional layer with a softmax activation to predict the probability score map $P^H$.  Given holistic depth bin centers and probability score map, we can now calculate the holistic depth map following Eq.~\ref{eq_depth}. 
Note that the ERP feature map is with the half resolution of the input ERP image to limit GPU memory usage.  
Therefore, we additionally employ an up-sampling module to upscale the probability score map to the desired resolution (i.e., $H \times W$).

\begin{table*}[!t]
    \centering
    \resizebox{0.98\textwidth}{!}{ 
    \begin{tabular}{c|c|c|c|c|c|c|c|c|c}
    \toprule
    Datasets&Method& Patch size$/$FoV &Abs Rel $\downarrow$& Sq Rel $\downarrow$&RMSE $\downarrow$ &RMSE(log) $\downarrow$ &$\delta_1$ $\uparrow$ & $\delta_2$ $\uparrow$ & $\delta_3$ $\uparrow$\\
    \midrule
    \multirow{7}*{Stanford2D3D} & FCRN~\cite{Laina2016DeeperDP}&$-/-$&0.1837&-&0.5774&-&0.7230&0.9207&0.9731\\
    &BiFuse with fusion~\cite{Wang2020BiFuseM3}&$-/-$&0.1209&-&0.4142&-&0.8660&0.9580&0.9860\\
   &UniFuse with fusion~\cite{Jiang2021UniFuseUF}&$-/-$&0.1114&-&0.3691&-&0.8711&0.9664&0.9882\\
    &OmniFusion (2-iter)~\cite{Li2022OmniFusion3M}&$256\times256$ $/$ 80$^\circ$&0.0950&0.0491&0.3474&0.1599&0.8988&0.9769&0.9924\\&PanoFormer*~\cite{Shen2022PanoFormerPT}&$-/-$& 0.1131 &0.0723 & 0.3557& 0.2454& 0.8808& 0.9623& 0.9855 \\
    \cmidrule{2-10}
    &HRDFuse,Ours &$128\times128$ $/$ 80$^\circ$&0.0984&0.0530&0.3452&0.1465&0.8941&0.9778&0.9923\\
    &HRDFuse,Ours &$256\times256$ $/$ 80$^\circ$&\textbf{\textred{0.0935}}&0.0508&\textbf{\textred{0.3106}}&\textbf{\textred{0.1422}}&\textbf{\textred{0.9140}}&\textbf{\textred{0.9798}}&\textbf{\textred{0.9927}}\\
    \bottomrule
    \toprule
    \multirow{7}*{Matterport3D} & FCRN~\cite{Laina2016DeeperDP}&$-/-$&0.2409&-&0.6704&-&0.7703&0.9714&0.9617\\
    &BiFuse with fusion~\cite{Wang2020BiFuseM3}&$-/-$&0.2048&-&0.6259&-&0.8452&0.9319&0.9632\\
   &UniFuse with fusion~\cite{Jiang2021UniFuseUF}&$-/-$&0.1063&-&0.4941&-&0.8897&0.9623&0.9831\\
     
    &OmniFusion (2-iter) *~\cite{Li2022OmniFusion3M} &$256\times256$ $/$ 80$^\circ$&0.1007&0.0969&0.4435&0.1664&0.9143&0.9666&0.9844\\&PanoFormer*~\cite{Shen2022PanoFormerPT}&$-/-$& 0.0904&0.0764& 0.4470&0.1650 & 0.8816& 0.9661& 0.9878\\
    \cmidrule{2-10}
    &HRDFuse,Ours &$128\times128$ $/$ 80$^\circ$&0.0967&0.0936&\textbf{\textred{0.4433}}&\textbf{\textred{0.1642}}&\textbf{\textred{0.9162}}&\textbf{\textred{0.9669}}&0.9844\\
    &HRDFuse,Ours &$256\times256$ $/$ 80$^\circ$&0.0981&0.0945&0.4466
&0.1656&0.9147&0.9666&0.9842\\    
    \bottomrule
    \toprule
    \multirow{8}*{3D60} & FCRN~\cite{Laina2016DeeperDP}&$-/-$&0.0699&0.2833&-&-&0.9532&0.9905&0.9966\\
     &Mapped Convolution~\cite{Eder2019MappedC}&$-/-$&0.0965& 0.0371 &0.2966& 0.1413& 0.9068& 0.9854& 0.9967\\
    &BiFuse with fusion~\cite{Wang2020BiFuseM3}&$-/-$&0.0615& -& 0.2440& - &0.9699 &0.9927 &0.9969\\
   &UniFuse with fusion~\cite{Jiang2021UniFuseUF}&$-/-$&0.0466& - &0.1968 &- &0.9835 &0.9965& 0.9987\\
   &ODE-CNN~\cite{Cheng2020OmnidirectionalDE}&$-/-$&0.0467& 0.0124& 0.1728& 0.0793& 0.9814 &0.9967 &0.9989\\
    &OmniFusion (2-iter)~\cite{Li2022OmniFusion3M}&$128\times128$ $/$ 80$^\circ$&0.0430 & 0.0114 & 0.1808 & 0.0735 & 0.9859 & 0.9969 & 0.9989\\
    
     \cmidrule{2-10}
    &HRDFuse,Ours &$128\times128$ $/$ 80$^\circ$&0.0363&0.0103&0.1565&0.0594&0.9888&\textbf{\textred{0.9974}}&0.9990\\
    &HRDFuse,Ours &$256\times256$ $/$ 80$^\circ$&\textbf{\textred{0.0358}}&\textbf{\textred{0.0100}}&\textbf{\textred{0.1555}}&\textbf{\textred{0.0592}}&\textbf{\textred{0.9894}}&0.9973&\textbf{\textred{0.9990}}\\
    \bottomrule
    \end{tabular}}
    \vspace{-8pt}
    \caption{Quantitative comparison with the SOTA methods. $*$ represents that the model is re-trained following the official setting. \textbf{\textred{red}} indicates that our method achieves the best performance.
    }
    \vspace{-13pt}
    \label{tab:comparison-to-soat}
\end{table*}

\noindent \textbf{Regional distribution-based depth prediction.} 
Compared with the ERP branch, predicting an ERP format depth map from TP patches based on corresponding regional depth distributions meets two critical difficulties: 1) accurate and smooth fusion of individual TP patches; 2) capturing the holistic information for the ERP format depth output. To address them, we utilize the spatially aligned index map $M$ from the SFA module and the holistic query embedding $T^H$ from the ERP branch (See Fig.~\ref{fig:cddc}(b)). We first follow the generic depth distribution classification to collect regional depth bin centers from the collection of TP feature maps $\left\{F^\mathrm{TP}_n\right\}$ and concatenate them to obtain the tensor $\mathbf{c}^R$ with the size $ B\times N$. Then, with the spatial guidance of index map $M$, we can obtain an ERP format bin center map $M_{c}$ from bin center vector set $\mathbf{c}^R$ as:
\begin{equation}
\setlength{\abovedisplayskip}{3pt}
\setlength{\belowdisplayskip}{3pt}
M_{c}(i,j) = \sum_{n=1}^{N} M(i,j)_{n} \cdot \mathbf{c}^R_n
\end{equation}
where $(i,j)$ is the pixel coordinate, and $n$ is the patch index. The bin center map $M_c$ represents the depth distribution of each pixel with aggregated regional structural information. 
Meanwhile, we concatenate and average a collection of processed regional tokens, which record the regional structural information of each individual TP patch, to a tensor $T^R$. Similarly, the index map $M$ then helps to aggregate the regional structure in $T^R$ to a regional feature map $M_{key}$. Next, with $M_{key}$ as the ``keymap'' and $T^H$ as the ``query'', we can predict the regional probability score map $P^R$ and further output the ERP format regional depth map $D^R$. Note that the query embedding $T^H$ from the ERP branch provides necessary and favorable holistic guidance.
\textit{Due to the page limit, more details, \eg, network architecture, can be found in Table. 1 of the suppl. material.}
\begin{table*}[t!]
    \centering
    \scalebox{0.75}{
        \begin{tabular}{l|c|c|c|c|c|c|c|c|c|c|c|c}
    \toprule
    ERP branch   &TP branch   &geometric fusion   &SFA   &CDDC & FPS& $\#$Params& Abs Rel $\downarrow$& Sq Rel $\downarrow$&RMSE $\downarrow$ & $\delta_1$ $\uparrow$ & $\delta_2$ $\uparrow$& $\delta_3$ $\uparrow$\\
    \midrule
    \checkmark &&& & & \textbf{7.88} & \textbf{33.57M}  &0.1028&0.0985 &0.4543 & 0.9086 & 0.9658&0.9841\\
    \midrule
   &\checkmark &\checkmark & & &  3.56 & 37.09M  & 0.1018&0.0982&	0.4492&	0.9104&	0.9662& 0.9842\\
    \midrule
    \checkmark&\checkmark &\checkmark & & &   2.82 & 70.66M  & 0.0986&0.0944 &0.4466 & 0.9141 & 0.9664&0.9843\\
    \midrule
    \checkmark&\checkmark & & \checkmark& &  6.21& 49.95M &0.0991 &0.0956 & 0.4479 &0.9132 & 0.9666&0.9843\\
    \midrule
    \checkmark&\checkmark & \checkmark&& \checkmark& 3.23&56.96M & 0.0978 &0.0940 &0.4458 & 0.9146&0.9666& 0.9841\\\midrule
    \checkmark&\checkmark & &\checkmark& \checkmark& 5.52&53.77M & \textbf{0.0967} &\textbf{0.0936} &\textbf{0.4433} & \textbf{0.9162}&\textbf{0.9669}& \textbf{0.9844}\\
    \bottomrule
    \end{tabular}}
    \vspace{-5pt}
    \caption{The ablation results for individual components. Both ERP and TP branch are trained with the depth distributions following~\cite{Bhat2021AdaBinsDE}.}
    \vspace{-12pt}
    \label{tab:ab_indcom}
\end{table*}

\vspace{-8pt}
\subsection{The Final Output and Loss Function}
\vspace{-5pt}
\label{loss}
To obtain the final depth map, we adaptively fuse the depth prediction $D^{H}$ from the holistic contextual branch and depth prediction $D^{R}$ from the regional structural branch, which can be formulated as follows:
\begin{equation}
\setlength{\abovedisplayskip}{2pt}
\setlength{\belowdisplayskip}{2pt}
\resizebox{0.38\hsize}{!}{$
    D = w_0  D^H + w_1  D^R,
    $}
\end{equation}

\noindent where $w_0$ and $w_1$ are learnable parameters and $w_0+w_1=1$ (superiority of adaptive weighting is shown in Table.~\ref{tab:ab_weight}).
Following previous works~\cite{Li2022OmniFusion3M,Jiang2021UniFuseUF}, we adopt BerHu loss~\cite{IroLaina2016DeeperDP} for pixel-wise depth supervision, denoted as $\mathcal{L}_{depth}$.
Furthermore, to encourage the holistic distribution to be consistent with all depth values in the ground truth depth map, we adopt the commonly used bi-directional Chamfer loss~\cite{HaoqiangFan2016APS} as the holistic distribution loss $\mathcal{L}_{H_{bin}}$. Therefore, the total loss $\mathcal{L}_{total}$ can be written as:
\begin{equation}
\setlength{\abovedisplayskip}{2pt}
\setlength{\belowdisplayskip}{2pt}
\resizebox{0.5\hsize}{!}{$
  \mathcal{L}_{total}=\mathcal{L}_{depth} + \lambda \mathcal{L}_{H_{bin}},$
  }
\end{equation}
where $\lambda$ is a weight factor and set to 0.1 for all experiments empirically~\cite{Bhat2021AdaBinsDE}. 

\vspace{-3pt}
\section{Experiments}

\noindent \textbf{Datasets and Metrics.}
We conduct experiments on three benchmark datasets: Stanford2D3D~\cite{Armeni2017Joint2D}, Matterport3D~\cite{Chang2017Matterport3DLF}, and 3D60~\cite{Zioulis2018OmniDepthDD}. Note that Stanford2D3D and Matterport3D are real-world datasets, while 3D60 is composed of two synthetic datasets (SUNCG~\cite{ShuranSong2016SemanticSC} and SceneNet~\cite{AnkurHanda2016SceneNetAA}) and two real-world datasets (Stanford2D3D and Matterport3D). However, there exists a issue in the 3D60 dataset, which is mentioned by UniFuse~\cite{Jiang2021UniFuseUF} that the problematic rendering may cause some problems with depth prediction.

Following previous works~\cite{Li2022OmniFusion3M,Wang2020BiFuseM3, Jiang2021UniFuseUF}, we evaluate our method with the standard metrics: Absolute Relative Error (Abs Rel), Squared Relative Error (Sq Rel), Root Mean Squared Error (RMSE), Root Mean Squared Logarithmic Error (RMSE (log)), as well as a percentage metric with a threshold $\delta_t$, where $t \in \{ 1.25^1,1.25^2,1.25^3\}$. 
\textit{Due to the lack of space, the details of datasets and metrics can be found in the suppl. material.}

\begin{table}[!t]
    \centering
    \resizebox{0.95\linewidth}{!}{ 
    \begin{tabular}{c|c|c|c|c|c|c}
    \toprule
    Number& Patch size$/$FoV &Abs Rel $\downarrow$& Sq Rel $\downarrow$&RMSE $\downarrow$&$\delta_1$ $\uparrow$ & $\delta_2$ $\uparrow$\\
    \midrule
    10& \multirow{4}*{128$\times$128 $/$ 80$^\circ$ }&0.0996&0.0965&0.4491&0.9130&0.9664\\
    18 & &\textbf{0.0967} &\textbf{0.0936} &\textbf{0.4433} &\textbf{0.9162} &0.9669\\
    26 & & 0.0978& 0.0945& 0.4444&0.9151 &\textbf{0.9670}\\
    46 && 0.1232& 0.1178& 0.4996& 0.8780& 0.9563\\
    \midrule
        10& \multirow{4}*{256$\times$256 $/$ 80$^\circ$ }&0.0976&0.0948&0.4447&0.9152&0.9668\\
    18 & &0.0981&0.0945&0.4466&0.9147&0.9666\\
    26 & &0.0974 &0.0953 &0.4478 &0.9147 &0.9662\\
    46 & &\textbf{0.0966} &\textbf{0.0938} &\textbf{ 0.4432} &\textbf{0.9168} &\textbf{0.9668}\\
    \bottomrule
    \end{tabular}}
    \vspace{-8pt}
    \caption{The ablation results for the number of TP patches.}
    \vspace{-15pt}
    \label{tab:ab_patchnum}
\end{table}

\noindent \textbf{Implementation Details.}
We implement our method using Pytorch and train it on a single NVIDIA 3090 GPU. We use ResNet-34~\cite{He2016DeepRL}, pre-trained on ImageNet~\cite{JiaDeng2009ImageNetAL}, as the encoder. Following~\cite{Li2022OmniFusion3M}, we use  Adam~\cite{kingma2014adam} optimizer with cosine annealing~\cite{IlyaLoshchilov2016SGDRSG} learning rate policy and set the initial learning rate to $10^{-4}$. The default TP patch number is $N=18$, and the batch size is $4$. Following~\cite{Li2022OmniFusion3M}, we train 80 epochs for Stanford2D3D~\cite{Armeni2017Joint2D} and 60 epochs for Matterport3D~\cite{Chang2017Matterport3DLF}, and 3D60~\cite{Zioulis2018OmniDepthDD}. The input images are augmented only by horizontal translation and horizontal flipping.

\vspace{-3pt}
\subsection{Comparison with the state-of-the-arts}
\vspace{-3pt}

In Table. \ref{tab:comparison-to-soat}, we compare our HRDFuse with the SOTA methods on three benchmark datasets. For a fair comparison, we do not discuss self-supervised methods~\cite{vaswani2017attention,Lai2021OLANETS3,Kong2022SelfsupervisedI3}. Note that OmniFusion did not provide the pre-trained models on the Matterport3D dataset, thus we re-trained them with the official hyper-parameters. PanoFormer did not provide the experiment details, \eg, epochs; thus for fair comparison, we re-trained the model with the same setting using the official hyper-parameters for the same epochs. For all the datasets, we show the results of the proposed HRDFuse with TP patch sizes of $128 \times 128$ and $256 \times 256$.

As shown in Table.~\ref{tab:comparison-to-soat}, our HRDFuse \textbf{performs favorably against} the SOTA methods~\cite{Li2022OmniFusion3M,Jiang2021UniFuseUF,Wang2020BiFuseM3,Zioulis2018OmniDepthDD,Laina2016DeeperDP} \textbf{by a significant margin on two of the three datasets.} Specifically, for the Stanford2D3D dataset, our HRDFuse with the patch size of $256 \times 256$ outperforms UniFuse~\cite{Jiang2021UniFuseUF} by 16.07\% (Abs Rel), 15.85\% (RMSE), and 4.29\% ($\delta_1$). Compared with OmniFusion (2-iter), our HRDFuse improves RMSE(log) by 11.07\% and $\delta_1$ by 1.52\%. \textit{More comparisons with it can be found in the suppl. material due to the space limit}. 

For Matterport3D and 3D60 datasets, which contain more samples, our HRDFuse is more advantageous and surpasses the compared methods for all metrics. 
On the Matterport3D dataset, our HRDFuse with the patch size $128 \times 128$ outperforms UniFuse by 2.65\% ($\delta_1$), 9.03\% (Abs Rel), outperforms PanoFormer by 3.46\% ($\delta_1$) and outperforms OmniFusion (2-iter) by 3.97\%(Abs Rel), 3.41\% (Sq Rel). On the 3D60 dataset, HRDFuse with the patch size $256 \times 256$ outperforms UniFuse by 23.18\% (Abs Rel) and 20.99\% (RMSE), and outperforms OmniFusion (2-iter) by 16.74\% (Abs Rel) and 13.99\% (RMSE). 

In Fig.~\ref{fig:compare}, we present the qualitative comparison with UniFuse~\ref{fig:compare}(b) and OmniFusion~\ref{fig:compare}(c). Our HRDFuse can recover more regional structural details (e.g., leaves and seats) and suffer less from artifacts caused by the discontinuity among TP patches (red boxes). 
\textit{More qualitative comparisons can be found in the suppl. material}. 

\begin{table}[!t]
    \centering
    \scalebox{0.8}{    
    \begin{tabular}{c|c|c|c|c}
    \toprule
    Patch FoV& Patch size &Abs Rel $\downarrow$& Sq Rel $\downarrow$&RMSE $\downarrow$\\
    \midrule
    \multirow{2}*{60} &128$\times$128&\textbf{0.0986}&0.0961&0.4454\\
    & 256$\times$256&	0.0986&	\textbf{0.0942}&\textbf{0.4448}\\
    \midrule
    \multirow{2}*{80} &128$\times$128& \textbf{0.0967}&	\textbf{0.0936}&	\textbf{0.4433} \\
    & 	256$\times$256& 0.0981&	0.0945&	0.4466	 \\
    \midrule
    \multirow{2}*{100} &128$\times$128&  \textbf{0.0970}	&\textbf{0.0938}&\textbf{0.4453}\\
    & 	256$\times$256&  0.0979&	0.0940&	0.4458\\
    \bottomrule
    \end{tabular}}

    \vspace{-5pt}
    \caption{The ablation results for the TP patch size and FoV.}
    \vspace{-15pt}
    \label{tab:ab_psize}
\end{table}

\vspace{-2pt}
\subsection{Ablation Study and Analyses}
\vspace{-2pt}
\label{abl_study}
\noindent \textbf{The effectiveness of each module.} We verify the effectiveness of each module in our HRDFuse by adding one module each time (Table.~\ref{tab:ab_indcom}).
We form our baselines in three ways. 
Firstly, for the ERP branch-only baseline, we directly follow the Adabins~\cite{Bhat2021AdaBinsDE} to predict the holistic depth distributions from the ERP images and regress the depth maps.
Secondly, with only the TP branch, we add the geometric fusion, as done in ~\cite{Li2022OmniFusion3M}, to the feature extractor to obtain the ERP format depth map. 
Thirdly, we combine the ERP branch and TP branch, followed by the geometric fusion mechanism in~\cite{Li2022OmniFusion3M}. Based on this, we then add the SFA module. Here, we directly leverage the spatially aligned index map to aggregate the patch feature vectors ${V_n}$ into an ERP feature map and predict the depth map, without employing the decoder (see Fig.~\ref{fig:overview}) or geometric fusion module of~\cite{Li2022OmniFusion3M} in the TP branch. Lastly, we add the CDDC module to learn the holistic-with-regional depth distributions.

\begin{table}[t!]
    \centering
    \footnotesize
    \resizebox{\linewidth}{!}{ 
    \begin{tabular}{c|c|c|c|c|c}
    \toprule
    Number of bins & Abs Rel $\downarrow$& Sq Rel $\downarrow$&RMSE $\downarrow$ & $\delta_1$ $\uparrow$ & $\delta_2$ $\uparrow$\\
    \midrule
    20  &0.0997&0.0963 & 0.4502& 0.9132&0.9661\\

    50& 0.0971 &0.0939 &0.4454 & 0.9159 & 0.9665 \\

    100&  \textbf{0.0967}& \textbf{0.0936}&\textbf{0.4433} &\textbf{0.9162} &\textbf{ 0.9669} \\
    150& 0.0997 & 0.0948 & 0.4497& 0.9121& 0.9662\\
    \bottomrule
    \end{tabular}}
    \vspace{-7pt}
    \caption{Impact of the number $B$ of depth histogram bins.}
    \vspace{-15pt}
    \label{tab:ab_binno}
\end{table}
As shown in Table.~\ref{tab:ab_indcom}, with the ERP branch alone, it is difficult to alleviate the projection distortion, thus leading to the worst depth estimation performance. The performance improves when using the TP branch only due to less distortion, and is further improved by the fusion of the ERP branch and TP branch (with the geometric fusion mechanism). Furthermore, by introducing the SFA module, the network parameters are significantly reduced by 29.31$\%$, leading to more than three frames per second (FPS) gain in inference speed. When the CDDC module is finally added, the performance is further boosted by 2.42\%(Abs Rel) and 2.09\%(Sq Rel), although the parameters slightly increase. Especially, compared with PanoFormer (20.37 M parameters), our method higher FPS (5.52) than it (4.93).

\noindent \textbf{Patch size, FoV, and the number of patches of TP.} They are essential parameters and directly affect the accuracy and efficiency of our method. Thus, we study their impact and find an optimal balance between efficiency and performance. Following~\cite{Li2022OmniFusion3M}, we fix the patch number as 18 and examine how TP patch size affects the learning under multiple patch FoVs. As in Table~\ref{tab:ab_psize}, on the Matterport3D dataset, all the results with the patch size of $128 \times 128$ perform better than those of $256 \times 256$, which indicates that too large patch size may cause the redundancy of regional structural information and degrade the accuracy of the final ERP output. Meanwhile, we can observe the influence of patch FoV in Table~\ref{tab:ab_psize}: either too small patch FoV or too large patch FoV degrades the performance. When FoV is too small, the regional information in each TP patch would be insufficient; in contrast, too large FoV increases the inconsistency in the overlapping areas between adjacent TP patches.

Furthermore, as the number of TP patches and the computational memory cost are directly related, we fix the patch size and FoV to compare the depth results with different patch numbers such that we can find the most cost-effective patch number. As shown in Table.~\ref{tab:ab_patchnum}, too few patches can not provide sufficient region-wise structural details, while too many patches lead to the redundancy of details, thus degrading the role of holistic contextual information.  We find that $N=18$ performs best in our experiments.

\noindent \textbf{Number of bins.} We now compare the performance with various numbers of depth distribution histogram bins. As observed from Table.~\ref{tab:ab_binno}, starting from $B=20$, the depth accuracy first improves with the increase of $B$, and then drops significantly. The result indicates that too many bins lead to difficulty in classification. For this reason, we choose $100$ as the number of bins for experiments.

\noindent \textbf{Weights of fusion.}  Table.~\ref{tab:ab_weight} lists the depth results under 4 groups of fusion weights with the patch number set as $N=18$, patch size as 128$\times$128, and FoV as 80$^\circ$. Overall, our adaptive weighting achieves the best performance.

\noindent \textbf{Rationality of SFA module.} As depicted in Fig.~\ref{fig:tp-erp} and Table.~\ref{tab:ab_indcom}, the geometric fusion of \cite{Li2022OmniFusion3M} requires more inference time to ensure the depth values of overlapping areas among TP patches. By contrast, our SFA module can provide the alignment in the feature space, which is more efficient and effective. As shown in \textit{Fig.~7 in the suppl. material}, when the holistic scene structure is simple, SFA module makes the index map (\textit{Fig.~7(b)}) centralized to several representative TP patches with higher frequency of index (\eg, index 4, 6 in \textit{Fig.~7(c)}) to avoid redundant usage. This indeed validates the overlap among the TP patches, as shown in Fig.~\ref{fig:tp-erp}. In comparison, when the scene structure becomes more complex (\textit{Fig.~8 in the suppl. material}), more TP patches (with index 12,16 in \textit{Fig.~8(c)}) are needed to describe the holistic depth information.


\begin{table}[t!]
    \centering
    \resizebox{\linewidth}{!}{ 
    \begin{tabular}{c|c|c|c|c|c|c}
    \toprule
     ERP branch& TP branch& Abs Rel $\downarrow$& Sq Rel $\downarrow$&RMSE $\downarrow$ & $\delta_1$ $\uparrow$ & $\delta_2$ $\uparrow$\\
    \midrule
     1& 0& 0.0976 & 0.0948& 0.4450&  0.9153& 0.9664\\

    0  & 1& 0.0975& 0.0944& 0.4459& 0.9149 & 0.9670 \\

    0.5  &0.5 &0.0969 & 0.0942 & 0.4442 & 0.9157&0.9668 \\
    \midrule
    \multicolumn{2}{c|}{Adaptive weighting} & \textbf{0.0967}& \textbf{0.0936}& \textbf{0.4433}& \textbf{0.9162}& \textbf{0.9669}\\
    \bottomrule
    \end{tabular}}
     \vspace{-7pt}
    \caption{ The ablation study for the final fusion.}
    \vspace{-16pt}
    \label{tab:ab_weight}
\end{table}

\vspace{-3pt}
\section{Conclusion and Future Work}
\vspace{-3pt}
This paper proposed a novel solution for monocular 360$^\circ$ depth estimation, which predicts an ERP format depth map by collaboratively learning the holistic-with-regional depth distributions. To address the two issues: 1) challenges in pixel-wise depth value regression; 2) boundary discontinuities brought by the geometric fusion, our HRDFuse introduced the SFA module and the CDDC module, whose contributions allow HRDFuse to efficiently incorporate ERP and TP, and significantly improve the depth prediction accuracy and obtain favorably better results.
Our work focused on the supervised monocular 360$^\circ$ depth estimation and did not cover self-supervised methods.
In the future, we will explore the potential of TP, e.g., contrastive learning for TP patches. In addition, our task and 360$^\circ$ semantic segmentation~\cite{Yang2021CapturingOC,Zhang2022BendingRD} are closely related, as they are both dense scene understanding tasks. Therefore, joint 360$^\circ$ monocular depth estimation and semantic segmentation  based on the combination of ERP and TP is a promising research direction.

\section*{Acknowledgement}

This work was supported by the CCF-Tencent Open Fund and the National Natural Science Foundation of China (NSFC) under Grant No. NSFC22FYT45. 

{\small
\bibliographystyle{ieee_fullname}
\bibliography{egbib}
}
\clearpage
\input{supp}

\end{document}

%% file: supp.tex
\newpage
\onecolumn 

\vspace{0.5in}
\begin{center}
 \rule{6.875in}{0.7pt}\\ 
 {\Large\bf Supplementary Material for\\ `` HRDFuse: Monocular 360$^\circ$ Depth Estimation by Collaboratively Learning Holistic-with-Regional Depth Distributions ''}
 \rule{6.875in}{0.7pt}
\end{center}
\appendix

\section{Abstract}
\label{ab:ab}
Due to the lack of space in the main paper, we provide more details of the proposed method and experimental results in the supplementary material. Sec.~\ref{ab:tp} adds more details of tangent projection. And in Sec.~\ref{ab:fs}, we further discuss the necessity of the SFA module. Sec.~\ref{ab:cddc} provides the detailed calculation progress of the Collaborative Depth Distribution Classification (CDDC) module. Sec.~\ref{ab:loss} introduces our loss function, and Sec.~\ref{ab:data} presents a detailed description of the used benchmark datasets and metrics. In the Sec.~\ref{ab:com_res}  and Sec.~\ref{ab:vis_res} , we show additional comparison results and visual results about experiments. Furthermore, we discuss the rationality of SFA module in the Sec.~\ref{ab:sfa} and show some comparison results on real data in the Sec.~\ref{ab:vis_real}.

\section{More Details of Tangent Projection}
\label{ab:tp}

\begin{figure*}[h]
    \centering
    \includegraphics[width=0.6\textwidth]{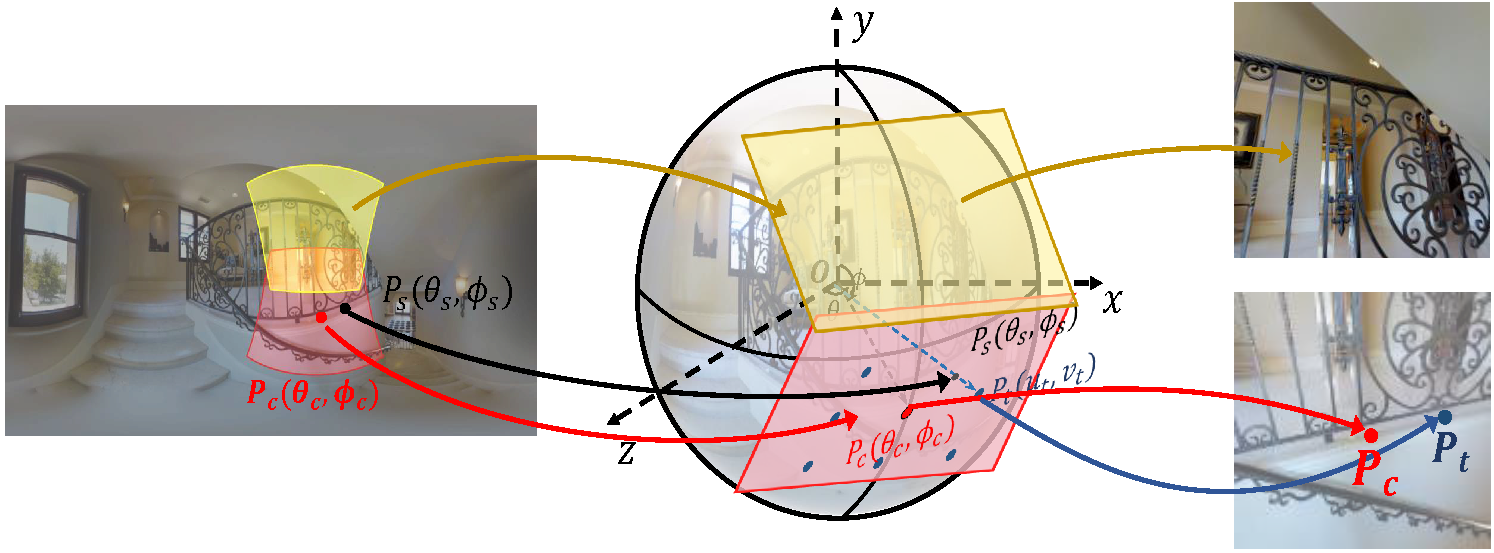}
    \vspace{-5pt}
    \caption{An example of TP and ERP. Two TP patches are projected from two different areas (red area and yellow area).}
    \label{fig:suptp}
    \vspace{-10pt}
\end{figure*}

We start by introducing an example of the tangent projection (TP)~\cite{Ai2022DeepLF}. As shown in Fig.~\ref{fig:suptp}, $P_s$ is a point on the sphere surface, $O$ is the center of the sphere, $P_c$ is the center of the tangent plane, and $P_t$ is the intersection point of the tangent plane and the extension line of $\overrightarrow{OP_s}$. As both $P_s$ and $P_c$ are on the sphere surface, we represent their spherical coordinates as $(\theta_s,\phi_s)$ and $(\theta_c,\phi_c)$, respectively. Then, we can obtain the planar coordinate $(u_t,v_t)$ of the point $P_t$ on the tangent plane as follows:
\begin{equation}
\small
\begin{split}
&    u_t=\frac{\cos(\phi_s)\sin(\theta_s-\theta_c)}{\cos(c)}, \\
&    v_t=\frac{\cos(\phi_c)\sin(\phi_s)-\sin(\phi_c)\cos(\phi_s)\cos(\theta_s-\theta_c)}{\cos(c)}, \\
&   \cos(c)=\sin(\phi_c)\sin(\phi_s)+\cos(\phi_c)\cos(\phi_s)\cos(\theta_s-\theta_c).
\end{split}
\label{eq.1}
\end{equation}
And the inverse transformations are:
\begin{equation}
\small
\begin{split}
& \theta_s = \theta_c + \tan^{-1}(\frac{u_t\sin(\sigma)}{\gamma\cos(\phi_c)\cos(\sigma)-v_t\sin(\phi_c)\sin(\sigma)}), \\
& \phi_s=\sin^{-1}(\cos(\sigma)\sin(\phi_c)+\frac{1}{\gamma}v_t\sin(\sigma)\cos(\phi_c)),
\end{split}
\label{eq.2}
\end{equation}

\noindent where $\gamma=\sqrt{u_{t}^{2}+v_{t}^{2}}$ and $\sigma=\tan^{-1}\gamma$. With Eq.\ref{eq.1} and Eq.~\ref{eq.2}, we can convert the points on the sphere and pixels in TP patches to each other. In addition, we can convert the spherical points into pixels in the ERP image with $(u_e,v_e)=(\frac{\theta_s*w}{2\pi},\frac{\phi_s*h}{\pi})$, where $w$ and $h$ are the width and height of the ERP image, respectively. Therefore, given the spherical coordinate of a TP patch center, we can achieve the mapping between the pixels in the ERP images and those in the corresponding TP patches. 

\begin{figure*}[h]
    \centering
    \includegraphics[width=0.8\textwidth]{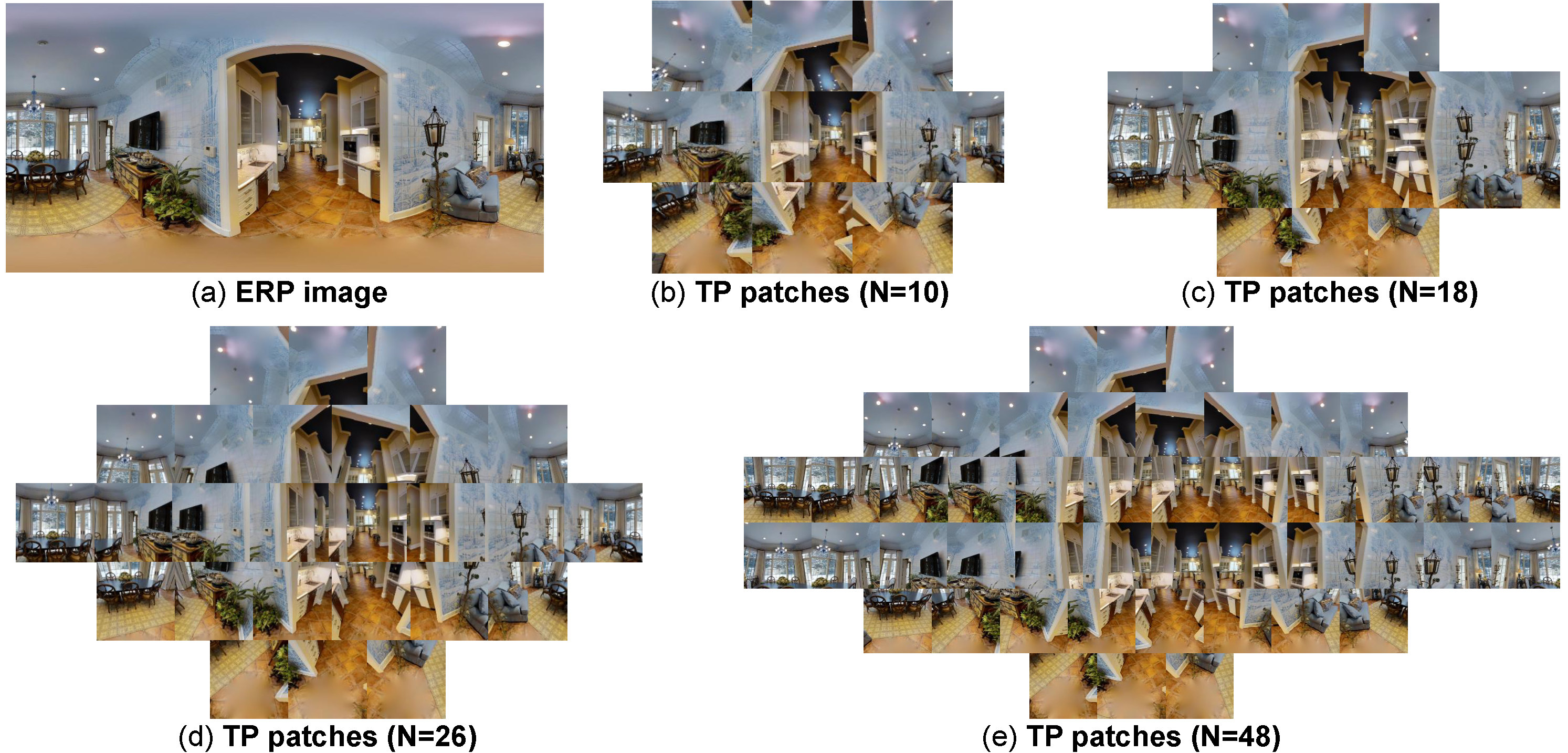}
    \vspace{-5pt}
    \caption{(a) An ERP image; (b) TP patches with the patch number $N=10$, which are sampled at three latitudes; (c) TP patches with $N=18$, which are sampled at four latitudes; (d) TP patches with $N=26$, which are sampled at five latitudes; (e) TP patches with $N=46$, which are sampled at six latitudes}
    \vspace{-10pt}
    \label{fig:tp_number}
\end{figure*}

\begin{figure*}[h]
    \centering
    \includegraphics[width=0.6\textwidth]{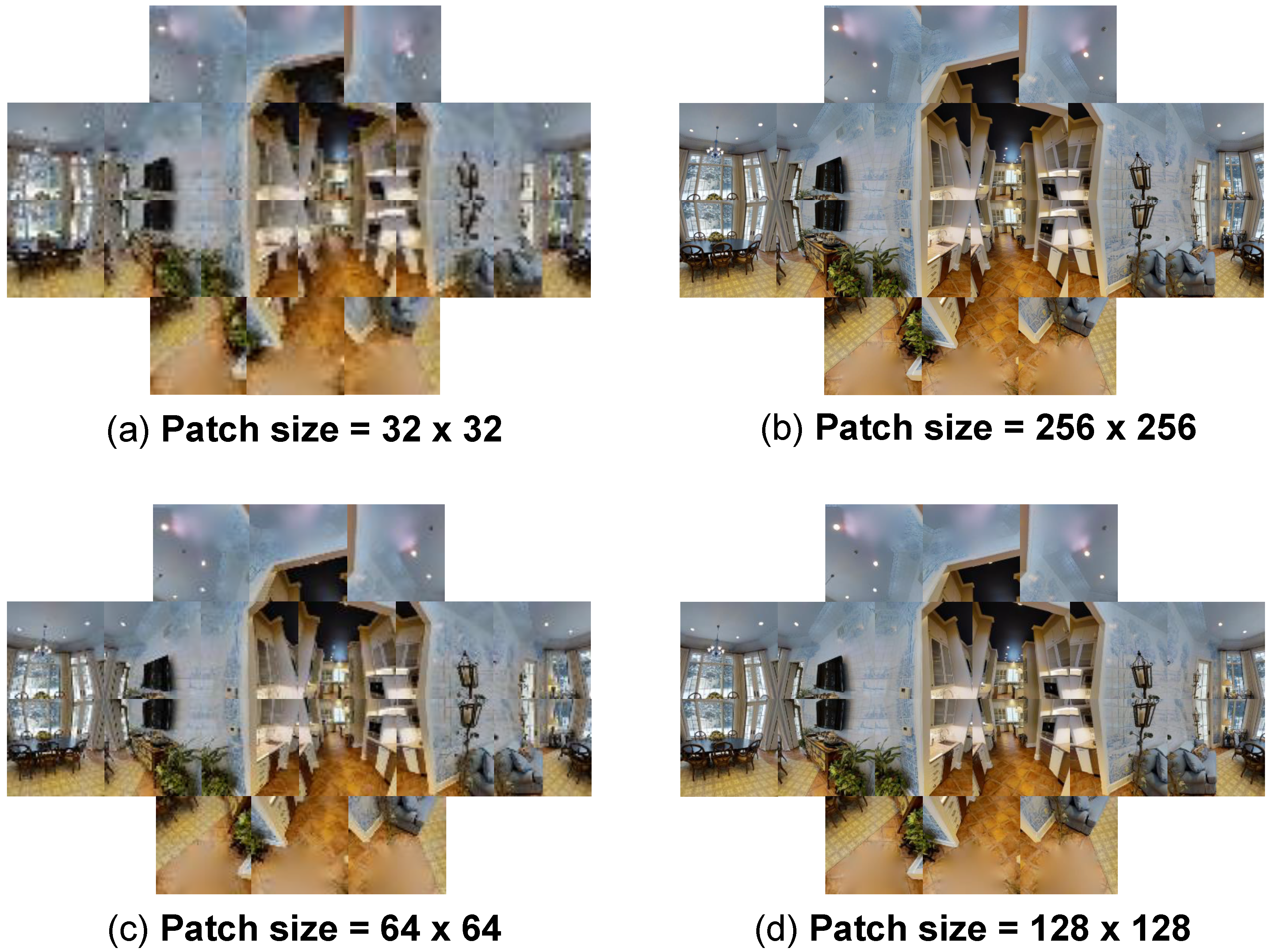}
    \vspace{-5pt}
    \caption{TP patches with different patch sizes.}
    \vspace{-5pt}
    \label{fig:tp_size}
\end{figure*}

\begin{figure*}[h]
    \centering
    \includegraphics[width=0.8\textwidth]{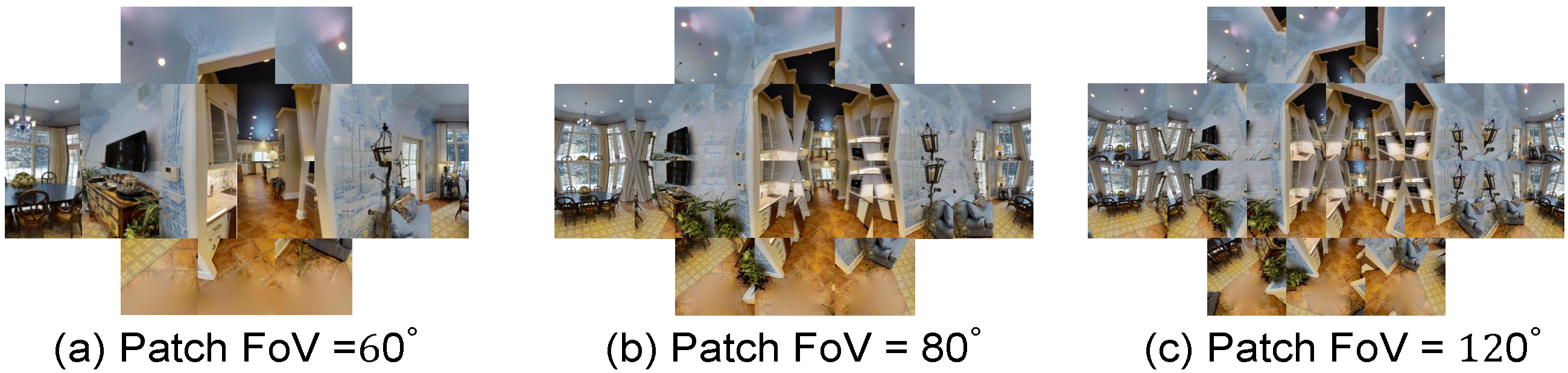}
    \vspace{-5pt}
    \caption{TP patches with different patch FoVs.}
    \vspace{-5pt}
    \label{fig:tp_fov}
\end{figure*}

The number of TP patches projected from a 360$^\circ$ spherical image depends on the sampling latitudes (the range of latitude is from -90$^\circ$ to 90$^\circ$) and the sampling number at each latitude.
For instance, in Omnifusion~\cite{Li2022OmniFusion3M}, TP patches are sampled from four latitudes: -67.5$^\circ$, -22.5$^\circ$, 22.5$^\circ$, 67.5$^\circ$, with 3, 6, 6, 3 patches on each latitude, respectively (see Fig~\ref{fig:tp_number}c). Besides, for one more case, as shown in  Fig~\ref{fig:tp_number}d, the sampled latitudes can be set: -72.2$^\circ$, -36.1$^\circ$, 0$^\circ$, 36.1$^\circ$, 72.2$^\circ$, while the sampled patch numbers are 3, 6, 8, 6, 3, respectively. From Fig~\ref{fig:tp_number}, we can see that with the patch number increased, the area of the overlapping regions increased correspondingly. As shown in Table.~\ref{fig:tp_number}, too few patches can not provide sufficient regional structural information, while too many patches lead to the redundancy of regional information. As a result, we chose to use a relatively small patch number of 18.

We fix the patch FoV to 80$^\circ$ and compare TP patches with different patch sizes of 32$\times$32, 64$\times$64, 128$\times$128, and 256$\times$256 in Fig.~\ref{fig:tp_size}, and it demonstrates that different patch sizes do not affect the content in each TP patch, but a large patch size does produce TP patches with more details. However, as shown in Table.~3 of the main paper, too large patch size will increase computational costs and the redundancy of regional structural information (the amount of pixels in the overlapping regions), which may further influence the prediction from holistic contextual information and decrease the overall performance. As a result, we chose to use a relatively large patch size of 128$\times$128.

For the patch FoV, we fix the patch size to 128$\times$128, and change the patch FoVs to obtain a set of TP patches, as shown in Fig.~\ref{fig:tp_fov}. Compared with the complete view of Fig.~\ref{fig:tp_number}a, too small FoV causes the loss of the scene information, while too large FoV causes the redundancy of information in the overlapping areas. As a result, we chose to use patch FoV $80^{\circ}$.

\section{More Discussion of Feature Similarity}
\label{ab:fs}
\begin{figure*}[h]
\vspace{-10pt}
    \centering
    \includegraphics[width=0.7\linewidth]{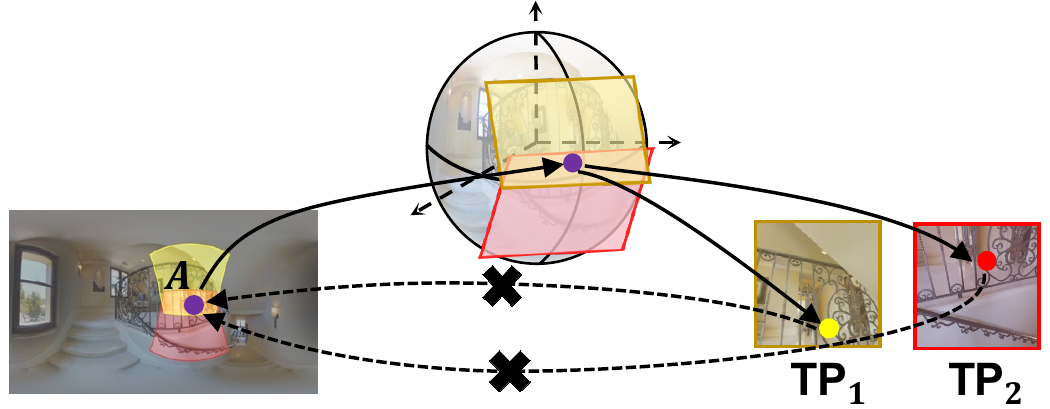}
    \vspace{-10pt}
    \caption{The relationships between pixels in ERP and TP patches.}
    \label{abl:fig1}
\vspace{-15pt}
\end{figure*}

Let us consider two pixels in two different TP patches but corresponding to the same pixel in the ERP image (See Fig.~\ref{abl:fig1}). Using a look-up table, we can easily search the locations in two different TP patches $\mathtt{TP}_1$ and $\mathtt{TP}_2$ \textit{w.r.t.} the pixel $A$. However, when re-projecting the pixel values in $\mathtt{TP}_1$ and $\mathtt{TP}_2$ to $A$ of ERP, we have to know which one is better (as done by OmniFusion), \textcolor{blue}{rendering the look-up table inapplicable for TP-to-ERP projection}. We will clarify this accordingly.

\section{More Details of Collaborative Depth Distribution Classification (CDDC)}
\label{ab:cddc}

In this section, we introduce the calculation process of the collaborative depth distribution classification (CDDC) module in detail.

First, given an ERP image with the size of $H_e\times W_e\times3$, we follow the gnomonic projection to obtain $N$ TP patches with the size of $H_t \times W_t \times 3$. Through the feature extractors, we can obtain the ERP feature map $f^E$ with the size $H_e/2 \times W_e/2 \times C_e$ and TP feature maps $\left\{f_n^T\right\}, n= 1, \dots, N$ with the size of $H_t/2 \times W_t/2 \times C_t \times N$, as the inputs of CDDC module. Then we summarize the detailed layer-by-layer network configurations in Table.~\ref{network}. Especially, we introduce network configurations in four parts: holistic depth distribution classification, holistic depth prediction, regional depth distribution classification, and regional depth prediction. 
\begin{table*}[t]
\centering
\scriptsize
\renewcommand\arraystretch{1.2}
\begin{tabular}{c|c|ccc|c|c|c}
\hline
\multicolumn{8}{c}{Collaborative Depth Distribution Classification (CDDC)} \\
\hline
  Input& InpRes & Kernel & Stride & Ch I/O  & Opt. & OutRes & Output  \\
\hline
\multicolumn{8}{c}{Holistic Depth Distribution Classification} \\
\hline
$F^{\mathrm{ERP}}$ & $H_e/2 \times W_e/2 \times C_e$ & 8 & 8 & $C_e/C_1$ & Flatten & $(\frac{H_e * W_e}{256}) \times C_1$ & $Tk^{H}_{in}$ \\
\hline
$Tk^{H}_{in}$ &  $(\frac{H_e * W_e}{256}) \times C_1$ &  - & -& $C_1/C_1$ & Transformer Encoder & $(\frac{H_e * W_e}{256}) \times C_1$  & $Tk^{H}_{out}$\\
\hline
$Tk^{H}_{out}[0]$   & $1 \times C_1$ & - & - & $C_1/B$ & Eq.~\ref{eq:sup1}, Eq.~\ref{eq:sup2} & $1 \times B$ & $\mathbf{c}^H$\\
\hline
\multicolumn{8}{c}{Holistic Range Attention Map} \\
\hline
$F^{\mathrm{ERP}}$& $H_e/2 \times W_e/2 \times C_e$ & 3 & 1 & $C_e/C_1$ &  & $H_e/2 \times W_e/2 \times C_1$ & $F^{H}$\\
\hline
$F^{H}\& $  &$H_e/2 \times W_e/2 \times C_1 \& $&- & -& - & \multirow{2}*{$\odot$}  & \multirow{2}*{$H_e/2 \times W_e/2 \times C_2$} & \multirow{2}*{$R^H$}\\
$Tk^{H}_{out}[1:C_2+1] (T^H)$  &$C_2\times C_1$&- & -& - & & & \\

\hline
$\mathcal{R}^H$& $H_e/2 \times W_e/2 \times C_2$  & - & - & - & Up-sample & $H_e \times W_e \times C_2$ & $\mathcal{R}^{H^{'}}$\\
\hline
$\mathcal{R}^{H^{'}}$& $H_e \times W_e \times C_2$  & 1 & 1 & $C_2/B$ & Softmax & $H_e \times W_e \times B$ & $P^H$\\
\hline
\multicolumn{8}{c}{Holistic Depth Prediction} \\
\hline
$\mathbf{c}^H \& P^H  $  & $1\times B\& H_e \times W_e \times B $ & - & - & $B/1$ & Eq.~\ref{eq:sup3} & $H_e \times W_e \times 1$ & $D^H$\\
\hline
\hline
\multicolumn{8}{c}{Regional Depth Distribution Classification} \\
\hline
$\left\{F_n^{\mathrm{TP}}\right\}$& $\frac{H_t}{2} \times \frac{ W_t}{2} \times C_t \times N$  & 4 & 4 & $C_t/C_1$ & Flatten &  $(\frac{H_t*W_t}{64}) \times C_1 \times N $ & $Tk^{R}_{in}$\\
\hline
$Tk^{R}_{in}$ & $(\frac{H_t* W_t}{64}) \times C_1 \times N $  & -& - & $C_1/C_1$ & Transformer Encoder  & $(\frac{H_t*W_t}{64}) \times C_1 \times N $ & $Tk^{R}_{out}$ \\
\hline
$Tk^{R}_{out}$[0]   &   $1 \times C_1 \times N$     &    -    &-& $C_1/B$ & Similar to Eq.~\ref{eq:sup1}, Eq.~\ref{eq:sup2}& $1 \times B \times N$ & $\mathbf{c}^R$\\
\hline
$\mathbf{c}^R \& M$    &   $1 \times B \times N \& \frac{H_e}{2}\times \frac{W_e}{2} \times N$  & - &-& - & Eq.~\ref{eq:sup4} & $\frac{H_e}{2}\times \frac{W_e}{2}\times B$ & $M_c$\\
\hline
\multicolumn{8}{c}{Regional Range Attention Map} \\
\hline
$Tk^{R}_{out}[1:C_2+1]$  &   $C_2 \times C_1 \times N$     &    -    &-& - & Mean & $C_1 \times N$ & $T^R$\\
\hline
$T^R \& M$   &   $C_1 \times N\& \frac{H_e}{2}\times \frac{W_e}{2} \times N$     &    -    &-& - & Similar to Eq.~\ref{eq:sup4} &  $\frac{H_e}{2}\times \frac{W_e}{2}\times C_1$ & $M_{key}$\\
\hline
$M_{key}\& T^H$   &   $\frac{H_e}{2}\times \frac{W_e}{2}\times C_1\& C_2 \times C_1$     &    -    &-& - &  $\odot$ &  $\frac{H_e}{2}\times \frac{W_e}{2}\times C_2$ & $R^R$\\
\hline
$R^R$&   $\frac{H_e}{2}\times \frac{W_e}{2}\times C_2$     &    -    &-& - &  Up-sample &  $H_e \times W_e\times C_2$ & $R^{R^{'}}$ \\
\hline
$R^{R^{'}}$&   $H_e \times W_e\times C_2$    &    1    &1& $C_2/B$ & Softmax &  $H_e \times W_e\times B$ & $P^R$ \\
\hline
\multicolumn{8}{c}{Regional Depth prediction} \\
\hline
$M_c$&  $\frac{H_e}{2}\times \frac{W_e}{2}\times B$   & -&-& - & Up-sample &  $H_e \times W_e\times B$ & $M_c^{'} $ \\
\hline
$M_c^{'} \& P^R $&   $H_e \times W_e \times B \& H_e \times W_e \times B $     &    -    &-& $B/1$ &  Similar to Eq.~\ref{eq:sup4} &  $H_e \times W_e\times1$ & $D^R$ \\
\hline
\end{tabular}
\caption{Network summary of the CDDC module ($\odot$ denotes the dot-production).}
\label{network}
\end{table*}

In the holistic depth distribution classification, given the output of the transformer encoder, embedding tokens $Tk^{H}_{out}$, we select the first token $Tk^{H}_{out}[0]$ to calculate the bin center vector $\mathbf{c}^H$ as
\begin{equation}
    \mathbf{c}^H_i = D_{min} +  (\mathbf{w}^H_i/2+\sum_{j=1}^{i-1}\mathbf{w}^H_j),
    \label{eq:sup1}
\end{equation}
\begin{equation}
    \mathbf{w}^H_i= (D_{max}-D_{min})\frac{(\mathrm{mlp}(Tk^{H}_{out}[0]))_i+ \epsilon}{\sum_{j=1}^{B}{(\mathrm{mlp}(Tk^{H}_{out}[0]))_j + \epsilon}},
    \label{eq:sup2}
\end{equation}
where $i,j=1,\dots,B$, $\mathbf{w}^H$ is the bin widths of the holistic distribution histogram,  $\mathrm{mlp}$ denotes a multi-layer perceptron (MLP) head with a ReLU activation, $(D_{min}, D_{max})$ is the depth range of the dataset, $B$ denotes the number of depth distribution bins,  and $\epsilon$ is a small constant to ensure that each value of $\mathbf{w}^H$ is positive. 
For the holistic depth prediction, the bin centers $\mathbf{c}^H$ are linearly blended with a probability score map $P^H$ to predict the depth value at each pixel $(i,j)$:
\begin{equation}
    D^{H}(i,j) = \sum_{b=1}^{B} P^{H}(i,j)_b \cdot \mathbf{c}^H_b.
\label{eq:sup3}
\end{equation}

For the regional depth distribution classification, as illustrated in the Table.~\ref{network}, we collect regional depth bin center vectors from the collection of TP feature map $\left\{F^\mathrm{TP}_n\right\}$ and concatenate the center vectors to obtain the tensor $\mathbf{c}^R$ with the size of $ B\times N$. Moreover, with the spatial guidance of index map $M$, we can obtain an ERP format bin center map $M_{c}$ based on $\mathbf{c}^R$ as follows:
\begin{equation}
    M_{c}(i,j) = \sum_{n=1}^{N} M(i,j)_{n} \cdot \mathbf{c}^R_n
\label{eq:sup4}
\end{equation}

where $(i,j)$ is the pixel coordinate, and $n$ is the patch index. The bin center map $M_c$ represents the depth distribution of each pixel with the regional structural information. Meanwhile, we concatenate the collection of selected tokens and reduce the first dimension of the concatenation with the average operation, to obtain the tensor $T^R$. Then we combine $T^R$ with the spatial locations of index map $M$ to obtain a feature map $M_{key}$. Moreover, we introduce the embedding vectors $T^H$ of the ERP branch. With $M_{key}$ as the ``keymap" and $T^H$ as the ``queries", we can predict the probability score map $P^R$ and further output the ERP format regional depth map $D^R$.

\begin{figure*}[h]
\centering
\includegraphics[width=0.7\linewidth]{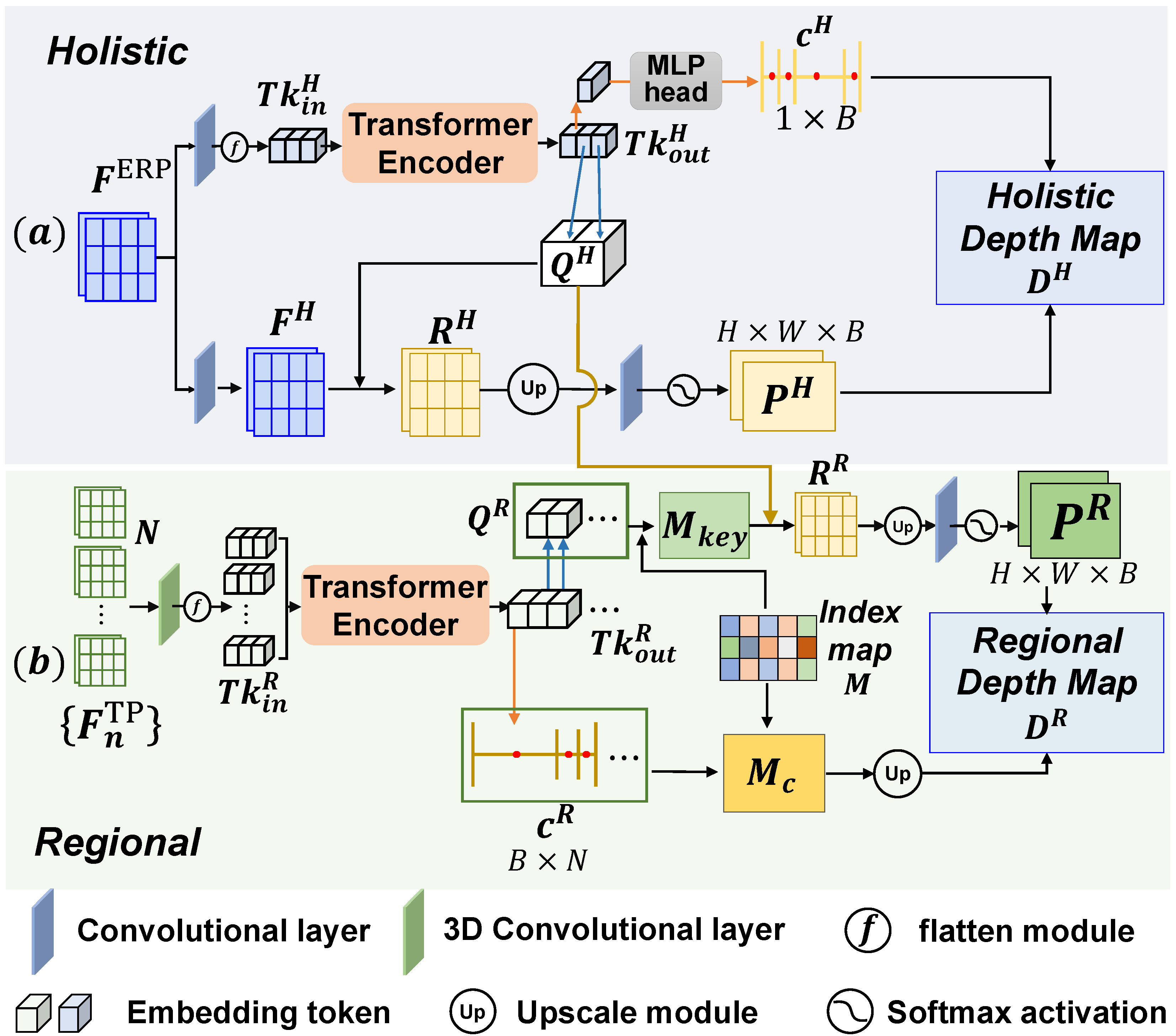}
\caption{Overview of the CDDC module with two steps.}
    \label{ab:fig:cddc}
\end{figure*}
\section{More Details of Loss Functions}
\label{ab:loss}
As introduced in the main paper, our loss consists of two terms: the pixel-wise depth loss and the holistic distribution loss. For the pixel-wise depth loss, following existing works~\cite{Li2022OmniFusion3M, Jiang2021UniFuseUF}, we adopt BerHu loss~\cite{IroLaina2016DeeperDP} for pixel-wise depth supervision, which is formulated as

\begin{align}
\small
&\mathcal{L}_{depth}=\sum_{i\in P}\mathcal{B}(D^i- D^i_{GT}),\\
    &\mathcal{B}(x)=\left\{\begin{aligned}	
		&\left | x\right |, \left | x\right |\leq c\\
		&\frac{x^2+c^2}{2c}, \left | x\right | > c\\
	\end{aligned}\right.
\end{align}

\noindent where $D_{GT}$ is the ERP format ground truth, $P$ indicates pixels which are valid in the ground truth depth map. c is a threshold hyper-parameter and set to 0.2 empirically~\cite{Li2022OmniFusion3M,Jiang2021UniFuseUF}.

Furthermore, following~\cite{Bhat2021AdaBinsDE}, we employ the bi-directional Chamfer Loss~\cite{HaoqiangFan2016APS} to encourage the holistic depth bin centers $\mathbf{c}^H(b)$ to be consistent with the distribution of all depth values (X) in the ground truth map as:
\begin{equation}
\small
  \mathcal{L}_{H_{bin}}= \textit{Cha}(X, \mathbf{c}^H(b))
\end{equation}
\begin{equation}
\small
  \textit{Cha}(X_1,X_2)=\sum_{x \in X_1} \min _{y \in X_{2}}\|x-y\|_{2}^{2}+\sum_{y \in X_{2}} \min _{x \in X_{1}}\|x-y\|_{2}^{2}
\end{equation}
Finally, the total loss is the summation of both two terms:
\begin{equation}
\small
  \mathcal{L}_{total}=\mathcal{L}_{depth} + \lambda \mathcal{L}_{H_{bin}}.
\end{equation}
For the balance weight $\lambda$, we follow~\cite{Bhat2021AdaBinsDE} and set $\lambda=0.1$ for all our experiments.

\section{More Details of Datasets and Metrics}
\label{ab:data}
We conduct experiments on three benchmark datasets: Stanford2D3D~\cite{Armeni2017Joint2D}, Matterport3D~\cite{Chang2017Matterport3DLF} and 3D60 dataset~\cite{Zioulis2018OmniDepthDD}. Note that Stanford2D3D dataset and Matterport3D dataset are real-world datasets, while 3D60 dataset is composed of two synthetic datasets (SunCG~\cite{ShuranSong2016SemanticSC} and SceneNet~\cite{AnkurHanda2016SceneNetAA}) and two real-world datasets (Stanford2D3D and Matterport3D). 
Stanford2D3D contains 1413 panoramic samples and we split it into 1,000 samples for training, 40 samples for validation and 373 samples for testing. Matterport3D is the largest real-world dataset for indoor panorama scenes containing 10,800 panoramas and we follow the official split to split it into 33875 samples for training, 800 samples for validation, and 1298 samples for testing. As the largest 360$^\circ$ depth estimation dataset, 3D60 totally contains 35973 panoramic samples where 33875 of them are used for training, 800 samples for validation, and 1298 samples for testing.  During training and testing, we resize the resolution of the panorama and depth map in the former two datasets into 512 $\times$ 1024. For 3D60, we set the input size into $256 \times 512$.

\section{Additional Comparison Results}
\label{ab:com_res}
As shown in the open source code of
PanoFormer~\cite{Shen2022PanoFormerPT}, the authors applied the masking strategy for the Stanford2D3D dataset:
\begin{lstlisting}[language=Python]
mask = torch.ones([512, 1024])
mask[0:int(512*0.15), :] = 0
mask[512-int(512*0.15):512, :] = 0
\end{lstlisting}

Therefore, we apply the same masking strategy
for the Stanford2D3D dataset and compare with PanoFormer in Table.~\ref{tab:sup_stan_comparison}. With the masking strategy, our HRDFuse outperforms PanoFormer~\cite{Shen2022PanoFormerPT} by a significant margin, \eg, 5.8$\%$ (Abs Rel), 11.3$\%$ (Sq Rel), 5.3$\%$ (RMSE). Furthermore, we compare our method with the PanoFormer on 3D60 dataset in Table.~\ref{tab:sup_3d60_comparison}. Note that PanoFormer did not provide the pre-trained models on the 3D60  dataset, we re-train the PanoFormer for 60 epochs with the official hyper-parameters and same experiment setting as OmniFusion and UniFuse. Our HRDFuse outperforms PanoFormer by a large margin.

\begin{table*}[h]
    \centering
    \resizebox{\textwidth}{!}{ 
    \begin{tabular}{c|c|c|c|c|c|c|c|c|c}
    \toprule
    Datasets&Method& Patch size$/$FoV &Abs Rel $\downarrow$& Sq Rel $\downarrow$&RMSE $\downarrow$ &RMSE(log) $\downarrow$ &$\delta_1$ $\uparrow$ & $\delta_2$ $\uparrow$ & $\delta_3$ $\uparrow$\\
    \midrule
    \multirow{6}*{Stanford2D3D} &PanoFormer*~\cite{Shen2022PanoFormerPT}&$-/-$& 0.1131 &0.0723 & 0.3557& 0.2454& 0.8808& 0.9623& 0.9855 \\
   &PanoFormer$^{\dag}$~\cite{Shen2022PanoFormerPT}&$-/-$& 0.0721 & 0.0506 & 0.3187& 0.1949& 0.9260& 0.9766& 0.9922 \\

    \cmidrule{2-10}
    &HRDFuse*,Ours &$128\times128$ $/$ 80$^\circ$&0.0984&0.0530&0.3452&0.1465&0.8941&0.9778&0.9923\\
    &HRDFuse$^\dag$,Ours &$128\times128$ $/$ 80$^\circ$&0.0730&0.0469& 0.3265&0.1311&0.9213&0.9807&0.9934\\
    &HRDFuse*,Ours &$256\times256$ $/$ 80$^\circ$&0.0935&0.0508&0.3106&0.1422&0.9140&0.9798&0.9927\\
    &HRDFuse$^\dag$,Ours &$256\times256$ $/$ 80$^\circ$&\textbf{\textred{0.0679}}&\textbf{\textred{0.0449}}&\textbf{\textred{0.3017}}&\textbf{\textred{0.1271}}&\textbf{\textred{0.9327}}&\textbf{\textred{0.9826}}&\textbf{\textred{0.9935}}\\
    \bottomrule
    \end{tabular}}
    \vspace{-8pt}
    \caption{Quantitative comparison with the SOTA methods. $*$ represents that the model is re-trained following the official setting. $\dag$ represents that the model is evaluated with the masking strategy in PanoFormer~\cite{Shen2022PanoFormerPT}.
    }
    \vspace{-10pt}
    \label{tab:sup_stan_comparison}
\end{table*}
\begin{table*}[h]
    \centering
    \resizebox{\textwidth}{!}{ 
    \begin{tabular}{c|c|c|c|c|c|c|c|c|c}
    \toprule
    Datasets&Method& Patch size$/$FoV &Abs Rel $\downarrow$& Sq Rel $\downarrow$&RMSE $\downarrow$ &RMSE(log) $\downarrow$ &$\delta_1$ $\uparrow$ & $\delta_2$ $\uparrow$ & $\delta_3$ $\uparrow$\\
    \midrule
    \multirow{10}*{3D60} & FCRN~\cite{Laina2016DeeperDP}&$-/-$&0.0699&0.2833&-&-&0.9532&0.9905&0.9966\\
     &RectNet~\cite{Zioulis2018OmniDepthDD}&$-/-$&0.0702&0.0297&0.2911&0.1017&0.9574&0.9933&0.9979\\
     &Mapped Convolution~\cite{Eder2019MappedC}&$-/-$&0.0965& 0.0371 &0.2966& 0.1413& 0.9068& 0.9854& 0.9967\\
    &BiFuse with fusion~\cite{Wang2020BiFuseM3}&$-/-$&0.0615& -& 0.2440& - &0.9699 &0.9927 &0.9969\\
   &UniFuse with fusion~\cite{Jiang2021UniFuseUF}&$-/-$&0.0466& - &0.1968 &- &0.9835 &0.9965& 0.9987\\
   &ODE-CNN~\cite{Cheng2020OmnidirectionalDE}&$-/-$&0.0467& 0.0124& 0.1728& 0.0793& 0.9814 &0.9967 &0.9989\\
    &OmniFusion (1-iter)~\cite{Li2022OmniFusion3M}&$128\times128$ $/$ 80$^\circ$&0.0469& 0.0127 &0.1880& 0.0792& 0.9827& 0.9963 &0.9988\\
    &OmniFusion (2-iter)~\cite{Li2022OmniFusion3M}&$128\times128$ $/$ 80$^\circ$&0.0430 & 0.0114 & 0.1808 & 0.0735 & 0.9859 & 0.9969 & 0.9989\\
    &PanoFormer*~\cite{Shen2022PanoFormerPT}&$-/-$& 0.0442 & 0.0124& 0.1691& 0.0676& 0.9861& 0.9966& 0.9987\\
     \cmidrule{2-10}
    &HRDFuse,Ours &$128\times128$ $/$ 80$^\circ$&0.0363&0.0103&0.1565&0.0594&0.9888&\textbf{\textred{0.9974}}&0.9990\\
    &HRDFuse,Ours &$256\times256$ $/$ 80$^\circ$&\textbf{\textred{0.0358}}&\textbf{\textred{0.0100}}&\textbf{\textred{0.1555}}&\textbf{\textred{0.0592}}&\textbf{\textred{0.9894}}&0.9973&\textbf{\textred{0.9990}}\\
    \bottomrule
    \end{tabular}}
    \vspace{-8pt}
    \caption{Quantitative comparison with the SOTA methods. $*$ represents that the model is re-trained following the official setting. 
    }
    \vspace{-10pt}
    \label{tab:sup_3d60_comparison}
\end{table*}

\begin{figure}[h]
\centering 
\begin{subfigure}[b]{0.9\textwidth}
        \includegraphics[width=\textwidth]{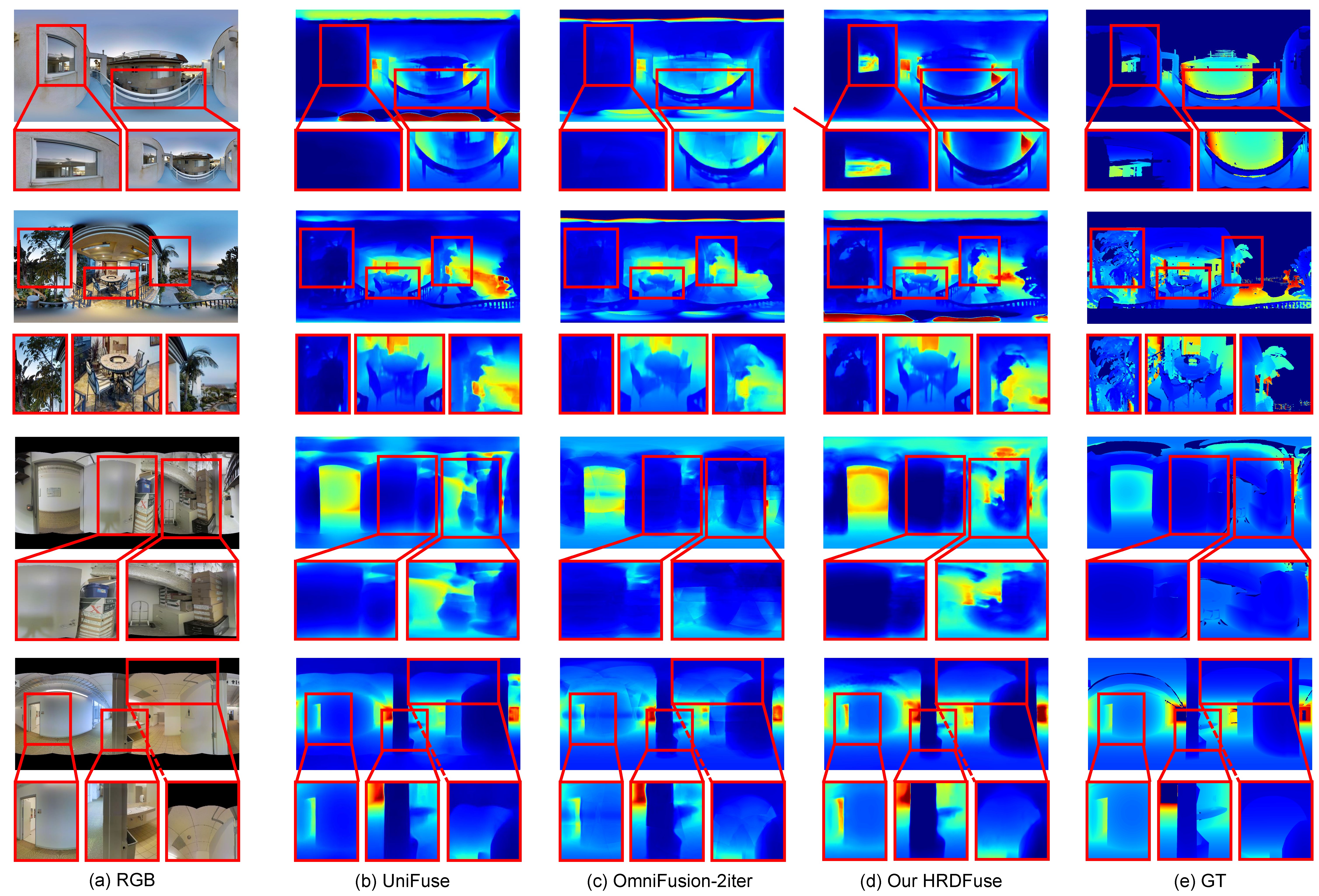}
        \caption{Visual comparisons on Stanford2D3D and Matterport3D.}
        \label{fig:supp_stan}
    \end{subfigure}%
    
~ 
    \begin{subfigure}[b]{0.9\textwidth}
        \includegraphics[width=\textwidth]{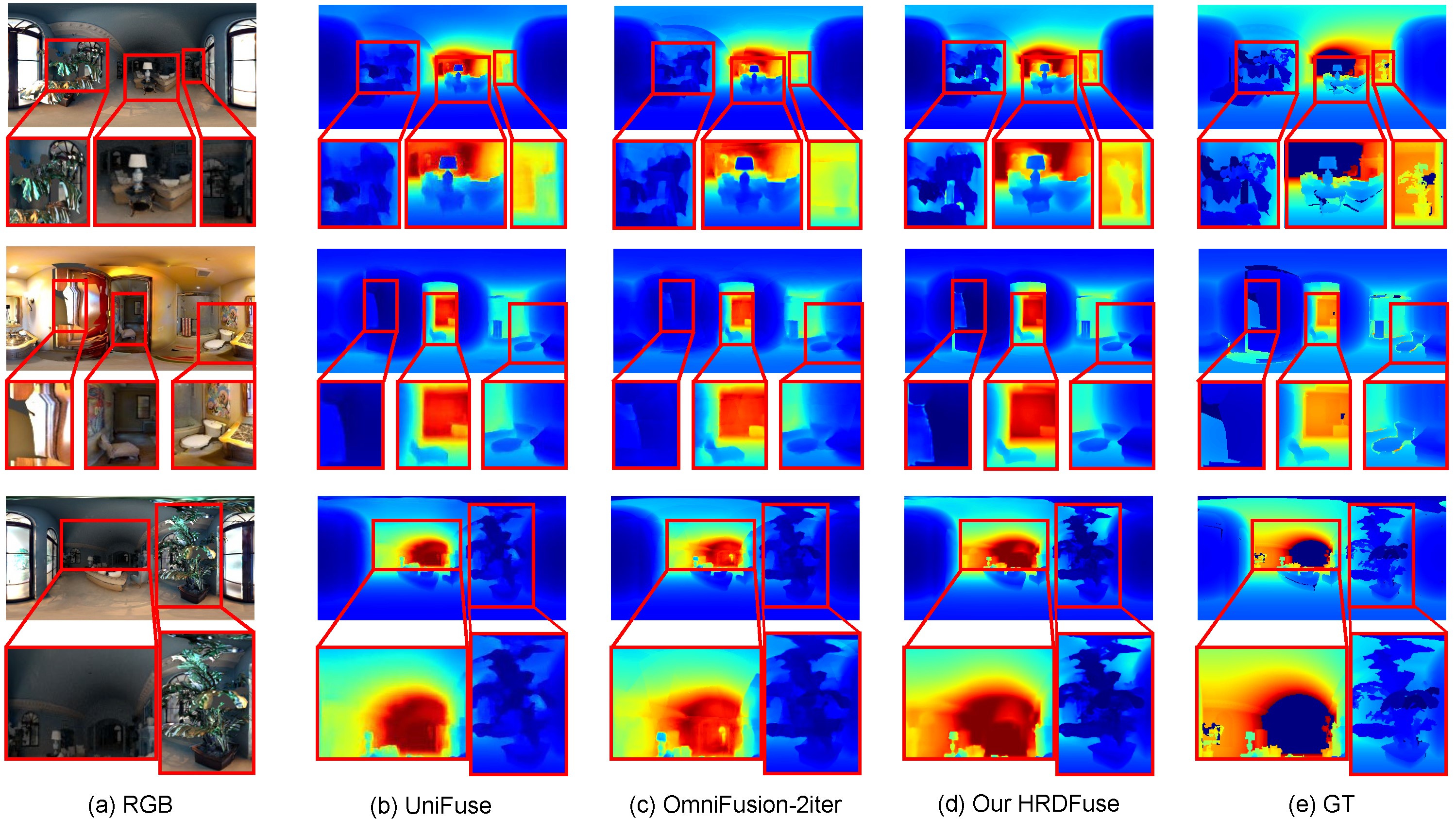}
        \caption{Visual comparisons on 3D60 dataset.}
        \label{fig:supp_3d60}
    \end{subfigure} 
 \label{fig:a}        
\caption{ More visual comparison results. }     
\label{fig}     
\end{figure}

\section{Additional Visual Results}
\label{ab:vis_res}

\noindent\textbf{More visual comparisons on Stanford2D3D and Matterport3D.} In Fig.~\ref{fig:supp_stan}, we perform qualitative comparisons with the SOTA methods, UniFuse.~\cite{Jiang2021UniFuseUF} and OmniFusion~\cite{Li2022OmniFusion3M}, on the Stanford2D3D dataset and Matterport3D dataset, whose samples are from real-world scenes. From the visual results, we confirm that our HRDFuse predicts the depth maps which are more precise and contain more structural details than other methods.

\noindent\textbf{More visual comparisons on 3D60.} In Fig.~\ref{fig:supp_3d60}, we perform qualitative comparisons with the SOTA methods, UniFuse.~\cite{Jiang2021UniFuseUF} and OmniFusion~\cite{Li2022OmniFusion3M}, on the 3D60 dataset, which contains both real-world and synthetic samples. From the visual results, we further confirm the superiority of our HRDFuse.

\section{Discussion on the rationality of SFA module}
\label{ab:sfa}

\begin{figure*}[h]
    \centering
\includegraphics[width=1\textwidth]{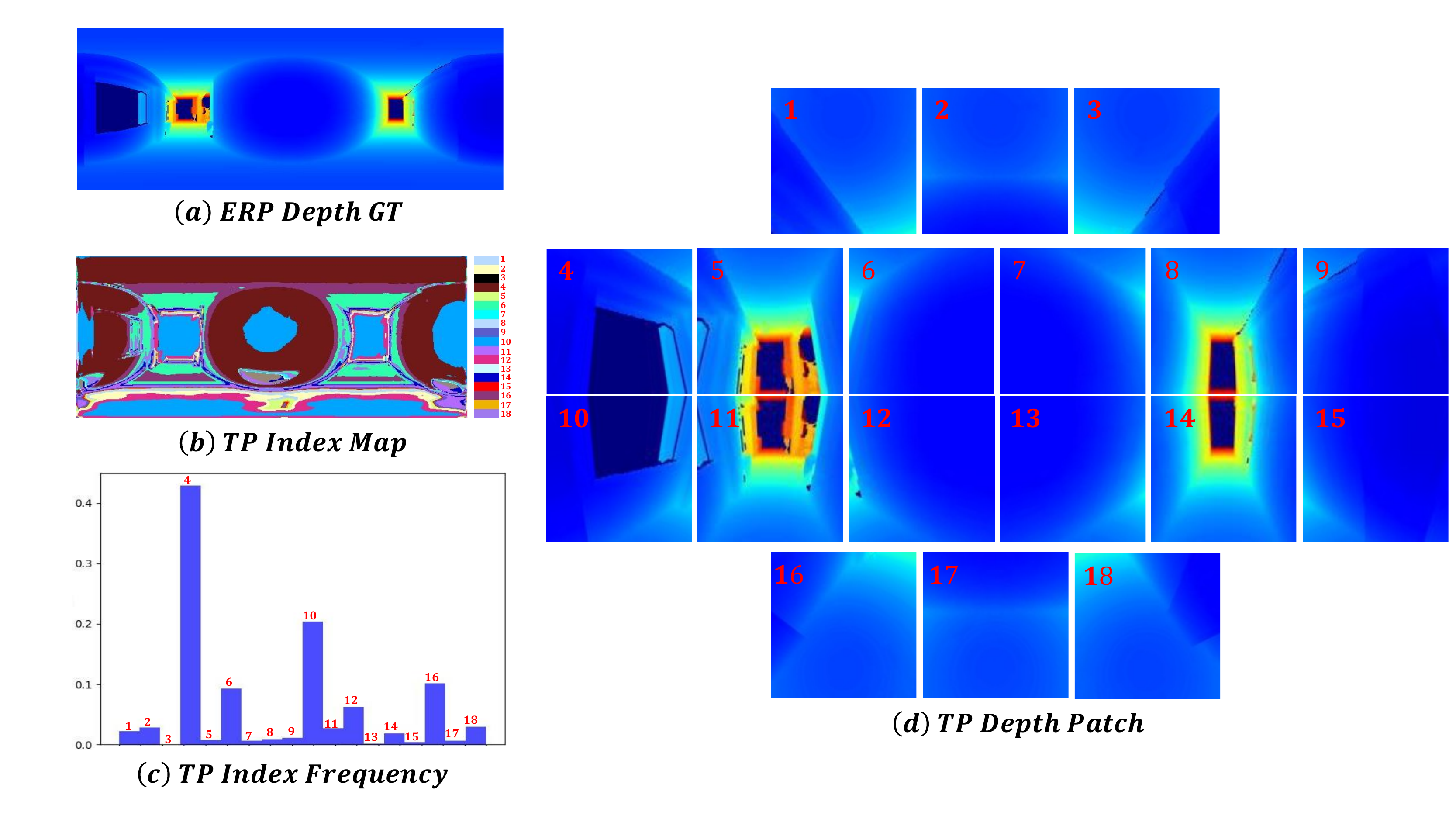}
    \caption{The visualization of (a) ERP depth ground truth, (b) TP index map (colored according to the attached color card), (c) TP index frequency and (d) TP depth patches with the corresponding index numbers from a scene with simple structure.}
    \label{fig:suppindex1}
\end{figure*}

\begin{figure*}[h]
    \centering
\includegraphics[width=1\textwidth]{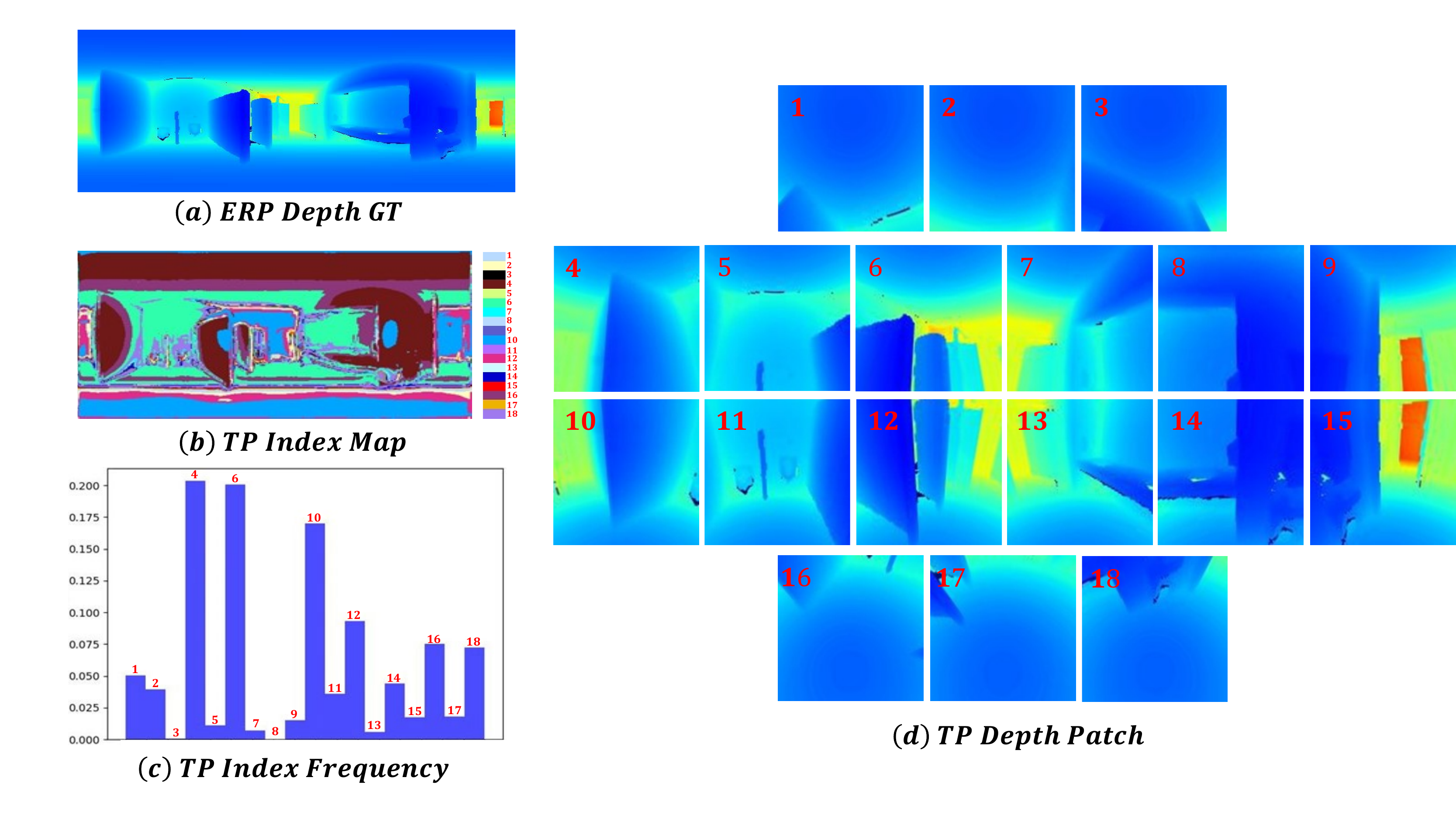}
    \caption{The visualization of (a) ERP depth ground truth, (b) TP index map (colored according to the attached color card), (c) TP index frequency and (d) TP depth patches with the corresponding index numbers from a scene with complex structure.}
    \label{fig:suppindex2}
\end{figure*}

As shown in the Fig~.\ref{fig:suppindex1}, for a scene with simple structure, our SFA module make the index map centralized to several representative TP patches with higher frequency of index, \eg, 4, 6, 10 (Fig.~\ref{fig:suppindex1}(c)). Combined with the Fig.~\ref{fig:suppindex1}(d), we can observe that the feature alignment based on the feature similarity in the SFA module tends to employ the most representative regional depth distributions to avoid the redundant usage. Meanwhile, facing the special depth values, SFA module will introduce the corresponding the regional depth distributions to predict them (\eg, with index 11, 14). Especially, with the scene structure becoming
more complex, the more TP patches are needed to describe the holistic depth information, as shown in the Fig.~\ref{fig:suppindex2}. The frequency of TP index is more balanced.
\clearpage
\section{Visual comparisons and discussion on real data.} 
\label{ab:vis_real}
\begin{figure*}[h]
    \centering
\includegraphics[width=1\textwidth]{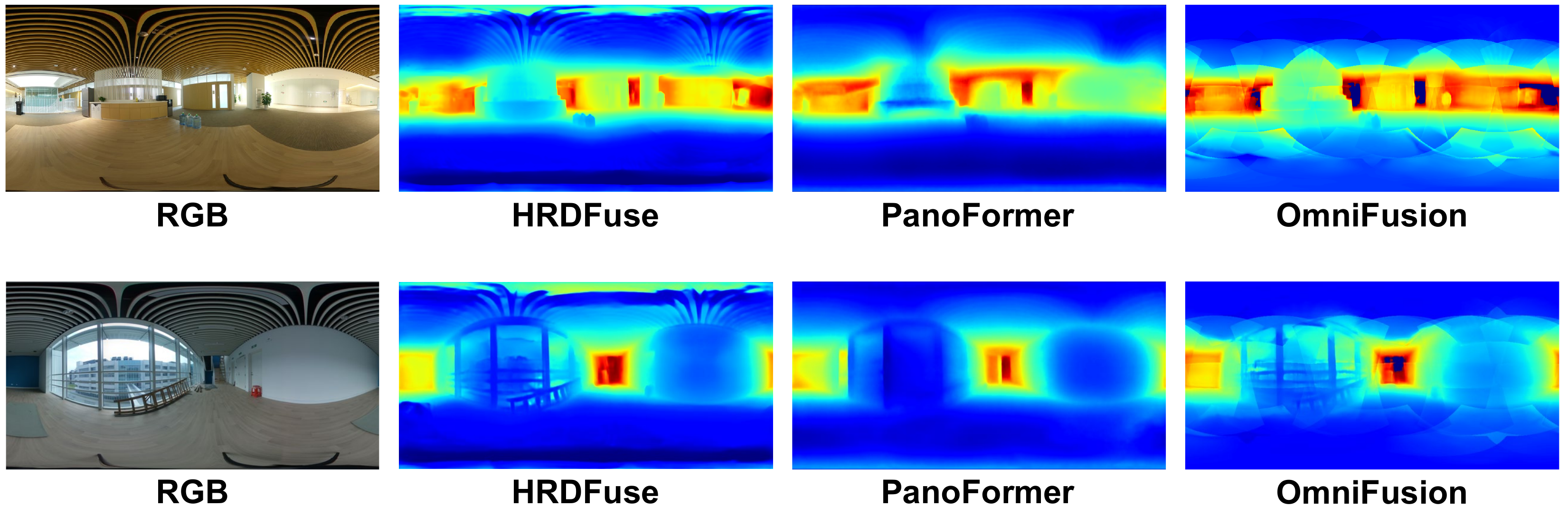}
    \caption{Visual comparisons on real data (captured by Ricoh Theta Z1).}
    \label{fig:suppreal}
\end{figure*}

To better compare the generation capability of our HRDFuse and other SoTA methods, we capture the two real images which records the indoor scene (considering the limited max depth value, we ignore the outdoor scene) and directly use the models trained on Matterport3D training dataset to predict their depth maps. As shown in the Fig.~\ref{fig:suppreal}, we can observe that our HRDFuse predicts more precise depth maps for the captured scenes. By contrast, the results of PanoFormer~\cite{Shen2022PanoFormerPT} tend to be blurry and over-smooth on unseen scenes.

\clearpage